\documentclass[runningheads]{llncs}
\usepackage{graphicx}
\usepackage{amsmath,amssymb} 
\usepackage{color}
\usepackage[width=122mm,left=12mm,paperwidth=146mm,height=193mm,top=12mm,paperheight=217mm]{geometry}

\usepackage{epsfig}
\usepackage{epstopdf}

\usepackage{paralist}

\usepackage{algorithm}
\usepackage{algorithmic}

\usepackage{times}
\begin{document}
\pagestyle{headings}
\mainmatter

\title{Recurrent Instance Segmentation} 

\titlerunning{Recurrent Instance Segmentation}

\authorrunning{Bernardino Romera-Paredes,\\
Philip Hilaire Sean Torr}

\author{Bernardino Romera-Paredes\\
Philip Hilaire Sean Torr}

\institute{Department of Engineering Science,\\
        University of Oxford\\
        \email{ bernard@robots.ox.ac.uk, philip.torr@eng.ox.ac.uk}
}



\maketitle

\begin{abstract}
Instance segmentation is the problem of detecting and delineating each distinct object of interest appearing in an image. Current instance segmentation approaches consist of ensembles of modules that are trained independently of each other, thus missing opportunities for joint learning. Here we propose a new instance segmentation paradigm consisting in an end-to-end method that learns how to segment instances sequentially. The model is based on a recurrent neural network that sequentially finds objects and their segmentations one at a time. This net is provided with a spatial memory that keeps track of what pixels have been explained and allows occlusion handling. In order to train the model we designed a principled loss function that accurately represents the properties of the instance segmentation problem. In the experiments carried out, we found that our method outperforms recent approaches on multiple person segmentation, and all state of the art approaches on the Plant Phenotyping dataset for leaf counting.
\end{abstract}

\section{Introduction}

Instance segmentation, the automatic delineation of different objects appearing
in an image, is a problem within computer vision that has attracted a fair
amount of attention. Such interest is motivated by both its potential
applicability to a whole range of scenarios, and the stimulating technical
challenges it poses.

Regarding the former, segmenting at the instance level is useful
for many tasks, ranging from allowing robots to segment a particular
object in order to grasp it, to highlighting and enhancing the outline of objects for the partially sighted, wearing ``smart specs'' \cite{smart}. Counting elements in an image has interest in
its own right \cite{arteta2013learning} as it has a wide range of
applications. For example, industrial processes that require the number
of elements produced, knowing the number of people who attended a
demonstration, and counting the number of infected and healthy blood cells in
a blood sample, required in some medical procedures such as malaria detection \cite{trager1976human}.

Instance segmentation is more challenging than other pixel-level learning problems such as semantic segmentation, which deals with classifying each pixel
of an image, given a set of classes. There, each pixel can belong to a set of predefined groups (or classes), whereas in instance segmentation the number of groups (instances) is 
unknown \textit{a priori}. This difference exacerbates the
problem:
where in semantic segmentation one can evaluate the prediction
pixel-wise, instance segmentation requires the clustering of pixels to be evaluated with a loss
function \emph{invariant to the permutation} of this assignment (i.e. it does
not matter if a group of ``person'' pixels is assigned to be person ``1'' or
person ``2''). That leads to further complexities in the learning of these
models. Hence, instance segmentation has remained a more
difficult problem to solve.

Most approaches proposed for instance level segmentation are based
on a pipeline of modules whose learning process is carried out independent
of each other. Some of them, such as \cite{hariharan2014simultaneous,liu2015multi},
rely on a module for object proposal, followed by another one implementing
object recognition and segmentation on the detected patches. A common problem with such piecewise learning methods is that each
module does not learn to accommodate itself to the outputs of other modules.
Another drawback is that in such cases, it
is often necessary to define an independent loss function for each module, which places
a burden on the practitioner to decide on convenient representations of the
data at intermediate stages of the pipeline.

These issues led us to take a different route and develop a fresh and
\emph{end-to-end} model able to learn the whole instance-segmentation process.
Due to the magnitude of the task, in this study we focus on the problem of
\emph{class-specific} instance segmentation. That is, we assume that the model
segments instances that belong to the \emph{same} class, excluding
classification stages. The solution to this problem is useful in its own right,
for example to segment and to count people in images \cite{vineet2011human}, but we consider that
this is also the first step towards general semantic learning systems that can
segment and classify different kinds of instances.
\begin{figure*}
\noindent \begin{centering}
\includegraphics[width=0.8\paperwidth]{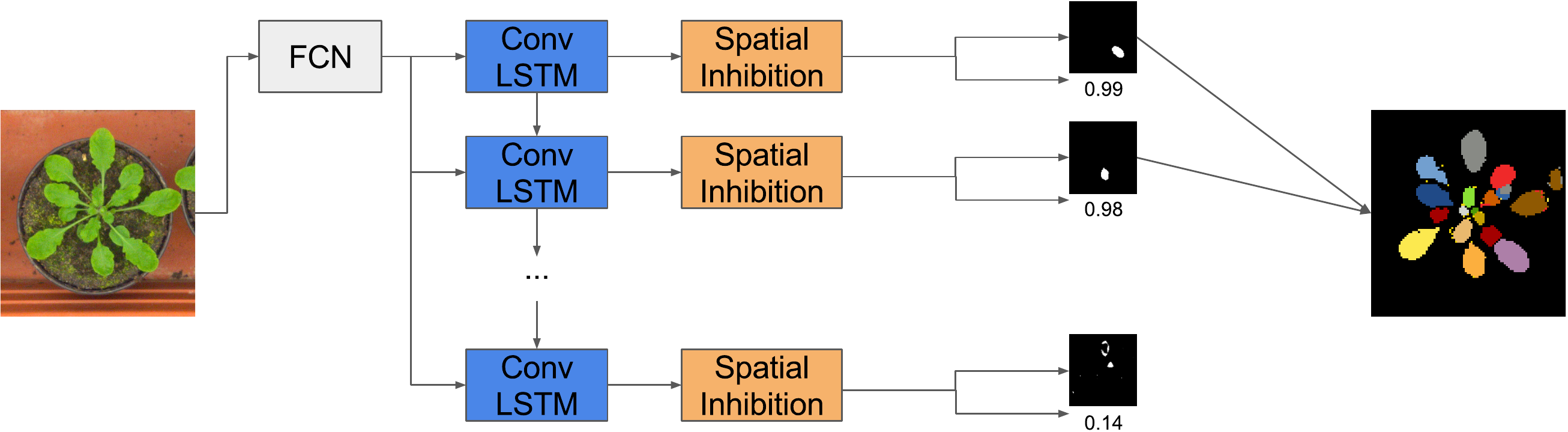}
\par\end{centering}

\caption{\label{fig:RIS} Diagram of Recurrent Instance Segmentation (RIS).}
\end{figure*}

The approach we propose here is partially inspired by how humans
count elements in a scene. It is known, \cite{dehaene1994dissociable,porter2007effort},
that humans count sequentially, using accurate spatial memory
in order to keep track of the accounted locations. Driven by this
insight, our purpose is to build a learning model capable of segmenting the
instances of an object in an image sequentially, keeping the current state in an
internal memory. In order to achieve this, we rely on recurrent neural networks
(RNNs), which exhibit the two properties discussed: the ability to produce
sequential output, and the ability to keep a state or memory along the sequence.

Our primary contributions, described in this paper, are
\begin{inparaenum}
\item the development of an \emph{end-to-end} approach for \emph{class specific}
  instance segmentation, schematized in Fig (\ref{fig:RIS}), based on RNNs containing convolutional layers, and
\item the derivation of a principled loss function for this problem.
\end{inparaenum}
%
We assess the capabilities of our model by conducting two experiments; one on
segmentation of multiple people, and the other on plant-leaves segmentation and
counting. 

\section{Background}

The work presented here combines several research areas. In this section
we summarize the main developments in these areas.

\subsection{Instance Segmentation Models\label{sub:instance_seg_review}}

Instance segmentation can be formulated as the conjunction
between semantic segmentation and object detection, for example in
\cite{ladicky2010and}, the authors proposed a model that integrates
information obtained at pixel, segment, and object levels. This is
because instance segmentation requires the capacity of object detection
approaches to separate between instances, and the ability of semantic
segmentation methods to produce pixel-wise predictions, and hence a
delineation of the shape of the objects.
The progress of instance segmentation methods is thus limited by the advances made in
both object detection and semantic segmentation.

A recent
breakthrough in object detection is the Region-based CNN (R-CNN) \cite{girshick2014rich}.
This approach consists in using a region proposal method to produce a large
set of varied sized object proposals from an image, then extracting
features for each of them by means of a CNN, and finally classifying
the resultant feature vectors. Several approaches for instance segmentation
build on this method. Two of them are \cite{hariharan2014simultaneous,liu2015multi},
which both use multiscale combinatorial grouping \cite{arbelaez2014multiscale}
as a region proposal method to extract candidates, followed by a region
refinement process. In the former work \cite{hariharan2014simultaneous},
the authors perform non-maximum suppression on the candidates, in
order to remove duplicates, and then they combine the coarse information
obtained by the CNN with superpixels extracted from the image in order
to segment the instances. In the latter work \cite{liu2015multi},
the output is refined by means of exemplar-based shape prediction
and graph-cut.
These approaches have produced state of the art results in instance
segmentation. However, they suffer a common drawback that we want
to avoid: they consist in an ensemble of modules that are trained
independently of each other.

Semantic segmentation methods have seen a significant
improvement recently, based first on the work in \cite{long2015fully}
which proposed a fully convolutional network that produces pixel-wise
predictions, followed by the works in \cite{chen2014semantic,Zheng_2015}
that improve the delineation of the predicted objects by using Conditional
Random Fields (CRFs). The work in \cite{liang2015proposal} builds
an instance segmentation model standing on these previous approaches.
This is based on predicting pixel-wise instances locations by a network, and then
applying a clustering method as a post-processing step, where the number of
clusters (instances) is predicted by another CNN. A problem with this approach is that
it is not optimizing a direct instance segmentation measure, but it relies on 
a surrogate loss function based on pixel distances to object positions.

One problem associated to instance segmentation is related to the
lack of an order among the instances in images (e.g. which instance
should be the first to be segmented). Several works \cite{tighe2014scene,yang2012layered,zhang2015monocular}
aim to overcome this problem by inferring and exploiting depth information
as a way to order the instances. They model that by means of a Markov
Random Field on the top of a CNN. One advantage of this approach is
that occlusion between objects is explicitly modeled. A disadvantage
is that this approach is not beneficial when instances appear at a
similar depth, for example, an image containing a sport team with
many players together.

Unlike previous approaches, we propose a new paradigm for instance
segmentation based on learning to segment instances sequentially,
\textit{letting the model decide} the order of the instances for each
image.

\subsection{Recurrent Neural Networks\label{sub:Recurrent-neural-networks}}

Recurrent neural networks (RNNs) are powerful learning models. Their
power resides in their capacity to keep a state or memory, in which
the model is able to store what it considers relevant events towards
minimizing a given loss function. They are also versatile, as they
can be applied to arbitrary input and output sequence sizes. These
two properties have led to the successful application of RNNs to a
wide range of tasks such as machine translation \cite{bahdanau2014neural},
handwriting recognition \cite{graves2009offline} and conversational
models \cite{vinyals2015neural}, among others.

RNNs are also useful for obtaining variable length information from
static images. One such example is DRAW \cite{gregor2015draw}, which
is a variational auto-encoder for learning how to generate images, in which
both the encoder and the decoder are RNNs, so that the image generation
process is sequential. Image captioning is another application which
has benefited from the use of RNNs on images. Examples of these are
\cite{donahue2015long,karpathy2014deep,vinyals2015show}, which are
based on using a CNN for obtaining
a meaningful representation of the image, which is then introduced
into an RNN that produces one word at each iteration. 
Another example, involving biological images, is in \cite{winther2015convolutional}, where the authors
explore a combination of convolutional layers and LSTMs in order to predict for each protein the 
subcellular compartment it belongs to.
In \cite{stewart2015end} the authors use an RNN to predict a bounding
box delimiting one human face at each iteration. This approach 
shares some of our motivations.
The main difference between both approaches is that in \cite{stewart2015end} they consider
a regression problem, producing at each iteration a set of scalars
that specify the bounding box where the instance is, whereas we consider
a pixel-wise classification problem. In \cite{pavel2015recurrent},
the authors present several recurrent structures to track and segment objects
in videos. Finally, it is worth mentioning the approach in \cite{williams2004greedy} despite not being an RNN, which consists in a greedy sequential algorithm for learning several objects in an image.

\subsection{Attention Based Models}

Attention based approaches consist in models that have the capability
of deciding at each time which part of the input to look at in order
to perform a task. They have recently shown impressive performance
in several tasks like image generation \cite{gregor2015draw}, object recognition \cite{ba2014multiple}, and
image caption generation \cite{xu2015show}. These approaches
can be divided into two main categories: hard, and soft attention
mechanisms. Hard attention mechanisms are those that decide which
part of the instance process at a time, totally ignoring the remainder \cite{mnih2014recurrent}.
On the contrary, soft attention mechanisms decide at each time a probability
distribution over the input, indicating the attention that each part
must receive. The latter are wholly differentiable, thus they can be optimized by backpropagation \cite{hermann2015teaching}.

Our approach resembles attention models in the sense that in both
cases the model selects different parts of the same input in successive
iterations. Attention based models are different from our approach
in which they apply attention mechanisms as a means to an end task
(e.g. image caption generation or machine translation), whereas in our
approach, attention to one instance at each time is the end target
we aim for.

\section{Segmenting One Instance at a Time\label{sec:model}}

In this section we describe the inference process that our approach
performs, as well as the structural elements that compose it.

The process at inference stage is depicted in Fig (\ref{fig:RIS}),
and can be described as follows: an image with height $h$ and width
$w$, $\mathbf{I}\in\mathbb{R}^{h\times w\times c}$ ($c$ is usually $3$,
or $4$ if including depth information) is taken as input of a fully
convolutional network, such as the one described in \cite{long2015fully}.
This is composed of a sequence of convolutional and max-pooling layers
that preserve the spatial information in the inner representations
of the image. The output of that network, $\mathbf{B}\in\mathbb{R}^{h'\times w'\times d}$,
represents the $d$-dimensional features extracted for each pixel, 
where the size of this map may be smaller than the size of the 
input image, $h'\leq h$, $w'\leq w$, due 
to the subsampling effect of the described network.
This output $\mathbf{B}$ will be the input to the RNN in all iterations in the
sequence. At the beginning of the sequence, the initial inner state
of the RNN, $\mathbf{h_{0}}$, is initialized to $0$. After the first iteration,
the RNN produces the segmentation of one of the instances in the image
(any of them), together with an indicator that informs about the confidence
of the prediction in order to have a stopping condition.  Simultaneously,
the RNN updates the inner state, $\mathbf{h_{1}}$, to account for the recent
segmented instance. Then, having again as inputs $\mathbf{B}$, and as inner
state $\mathbf{h_{1}}$, the model outputs another segmented instance and its confidence score.
This process keeps iterating until the confidence score drops below a certain level in 
which the model stops, ideally having segmented all instances in the image.

The sequential nature of our model allows to deal with common instance segmentation problems.
In particular it can implicitly model occlusion,
as it can segment non-occluded instances first, and keep in its state
the regions of the image that have already been segmented in order
to detect occluded objects. Another purpose of the state
is to allow the model to consider potential relationships from different
instances in the image. For example, if in the first iteration
an instance of a person embracing something is segmented, and this information is somehow
kept in the state, then in subsequent iterations, it might increase the plausibility of having
another person being embraced by the first one.

The structure of our approach is composed of a fully convolutional
network, followed by an RNN, a function to transform
the state of the RNN into the segmentation of an instance and its confidence score, and finally
the loss function that evaluates the quality of the predictions and that
we aim to optimize. The first of these components, the fully convolutional
network, is used in a similar manner as explained in \cite{long2015fully}.
In the following we describe in detail the remaining three components.

\subsection{Convolutional LSTM}

Long short-term memory (LSTM) networks \cite{hochreiter1997long}
have stood out over other recurrent structures because they are able
to prevent the vanishing gradient problem. Indeed, they are the chosen
model in most works reviewed in Sec. \ref{sub:Recurrent-neural-networks}
in which they have achieved outstanding results. In this section we
build on top of the LSTM unit, but we perform some changes in its structure
to adapt it to the characteristics of our problem.

In the problem we have described, we observe that the input to the
model is a map from a lattice (in particular, an image), and the output
is also a map from a lattice. Problems with these characteristics,
such as semantic segmentation \cite{long2015fully,Zheng_2015} and
optical flow \cite{fischer2015flownet}, are often tackled using structures
based on convolutions, in which the intermediate representations of
the images preserve the spatial information. In our problem, we can
see the inner state of recurrent units as a map that preserves
spatial information as well. That led us to convolutional versions
of RNNs, and in particular to convolutional long
short-term memory (ConvLSTM) units.

A ConvLSTM unit is similar to an LSTM one, the only difference
being that the fully connected layers in each gate are replaced by convolutions,
as specified by the following update equations:
\begin{equation}
\begin{array}{c}
\mathbf{i_{t}}={\rm Sigmoid}\left({\rm Conv}\left(\mathbf{x_{t}};\mathbf{w_{xi}}\right)+{\rm Conv}\left(\mathbf{h_{t-1}};\mathbf{w_{hi}}\right)+\mathbf{b_{i}}\right)\\
\mathbf{f_{t}}={\rm Sigmoid}\left({\rm Conv}\left(\mathbf{x_{t}};\mathbf{w_{xf}}\right)+{\rm Conv}\left(\mathbf{h_{t-1}};\mathbf{w_{hf}}\right)+\mathbf{b_{f}}\right)\\
\mathbf{o_{t}}={\rm Sigmoid}\left({\rm Conv}\left(\mathbf{x_{t}};\mathbf{w_{xo}}\right)+{\rm Conv}\left(\mathbf{h_{t-1}};\mathbf{w_{ho}}\right)+\mathbf{b_{o}}\right)\\
\mathbf{g_{t}}={\rm Tanh\,\,\,\,}\left({\rm Conv}\left(\mathbf{x_{t}};\mathbf{w_{xg}}\right)+{\rm Conv}\left(\mathbf{h_{t-1}};\mathbf{w_{hg}}\right)+\mathbf{b_{g}}\right)\\
\mathbf{c_{t}}=\mathbf{f_{t}}\odot \mathbf{c_{t-1}}+\mathbf{i_{t}}\odot \mathbf{g_{t}}\\
\mathbf{h_{t}}=\mathbf{o_{t}}\odot{\rm Tanh}(\mathbf{c_{t}})
\end{array},\label{eq:convlstm}
\end{equation}
where $\odot$ represents the element-wise product operator, $\mathbf{i_{t}}, \mathbf{f_{t}}, \mathbf{o_{t}}, \mathbf{g_{t}} \in\mathbb{R}^{h'\times w'\times d}$ are the gates, and $\mathbf{h_{t}},\mathbf{c_{t}}\in\mathbb{R}^{h'\times w'\times d}$ represents the memory
of the recurrent unit, being $d$ the amount of memory used for each pixel (in this paper we assume that the number of channels in the recurrent unit is the same as the number of channels produced by the previous FCN). Each of the filter weights ($\mathbf{w}$ terms) has dimensionality $d \times d \times f \times f$, where $f$ is the size of the filter, and each of the bias terms ($\mathbf{b}$ terms) is a $d$-dimensional vector repeated across height and width. We refer to the
diagrams and definitions presented in \cite{hochreiter1997long} for a detailed explanation
of the LSTM update equations, from which eq. (\ref{eq:convlstm}) follows. We also provide a diagram illustrating eq. (\ref{eq:convlstm}) in the Appendix Sec. B.
Note that a primary aspect of keeping a memory
or state is to allow our model to keep account of the pixels of the
image that have already been segmented in previous iterations of the
process. This can be naturally done by applying convolutions 
to the state. We can also stack two or more ConvLSTM units with the aim 
of learning more complex relationships.

The advantages of ConvLSTM with respect to regular LSTM go hand in
hand with the advantages of convolutional layers with respect to linear
layers: they are suitable for learning filters, useful for spatially
invariant inputs such as images, and they require less memory
for the parameters. In fact the memory required is independent of
the size of the input. A similar recurrent unit has been recently proposed
in \cite{shi2015convolutional} in the context of weather forecasting
in a region.

\subsection{Attention by Spatial Inhibition}

\begin{figure*}
\noindent \begin{centering}
\includegraphics[width=0.8\paperwidth]{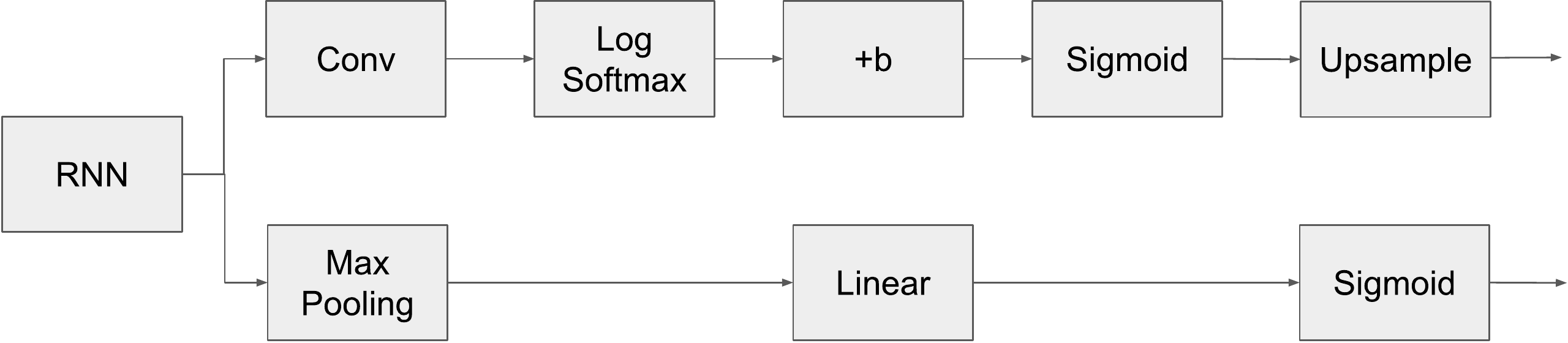}
\par\end{centering}

\caption{\label{fig:Inhibition} Diagram of the spatial inhibition module.}
\end{figure*}

The output produced by our model at time $t$ is a function $r(\cdot)$
of the hidden state, $\mathbf{h_{t}}$.
This function produces two outputs, $r(\cdot)\,:\,\mathbb{R}^{h'\times w'\times d}\rightarrow\left\{ [0,1]^{h\times w},[0,1]\right\} $.
The first output is a map that indicates which pixels compose the
object that is segmented in the current iteration. The second output
is the estimated probability that the current segmented candidate
is an object. We use this output as a stopping condition.
In the following we describe these two functions, supporting our presentation
on a schematic view of $r(\cdot)$, which is given in Fig. (\ref{fig:Inhibition}).

The function that produces the first output can be described as a
sequence of layers which have the aim of discriminating one, and only
one instance, filtering out everything else. Firstly we use a convolutional layer which maps the $d$ channels of the hidden state to $1$ output channel, using $1\times 1$ filters. This is followed by a log-softmax layer,
$f_{{\rm LSM}}(\mathbf{x})_{i}={\rm log}\left(\frac{{\rm exp}(x_{i})}{\sum_{j}{\rm exp}(x_{j})}\right)$,
which normalizes the input across all pixels, and then applies a logarithm.
As a result, each pixel value can be in the interval $(-\infty,0]$,
where the sum of the exponentiation of all values is $1$. This leads to
a competing mechanism that has the potential of inhibiting pixels
that do not belong to the current instance being segmented. Following
that, we use a layer which adds a learned bias term to the input data.
The purpose of this layer is to learn a threshold, $b$, which filters
the pixels that will be selected for the present instance. Then,
a sigmoid transformation is applied pixel-wise. 
Hence, the resultant pixel values are all in
the interval $[0,1]$, as required. Finally, we upsample the resultant $h' \times w'$ map
back to the original size of the input image, $h \times w$. 
In order to help understand the effect of these layers, we visualize in Sec. A in the Appendix, the inner representations captured by the model at different stages of the described pipeline.

The function which encodes the relationship between the current state
$\mathbf{h_{t}}$ and the confidence of the predicted candidate consists simply
of a max-pooling and a linear layer, followed by a sigmoid function.

\subsection{Loss Function\label{sub:Loss-Function}}

Choosing a loss function that accurately reflects
the objective we want to achieve is key for any model to be able to
learn a given task. In order to present the loss function that we
use, let us first add some notation. 

At training stage we are provided with the training set composed of
labeled images. We denote image $i$ as $\mathbf{I^{(i)}}\in\mathbb{R}^{h\times w\times c}$,
where for simplicity we consider the same size ($h\times w\times c$)
for all images. Its annotation $\mathbf{Y^{(i)}}=\left\{ \mathbf{Y_{1}^{(i)}},\mathbf{Y_{2}^{(i)}},\ldots,\mathbf{Y_{n_{i}}^{(i)}}\right\} $,
is a set of $n_{i}$ masks, $\mathbf{Y_{t}^{(i)}}\in\{0,1\}^{h\times w}$,
for $t\in\{1,\ldots,n_{i}\}$, containing the segmentation of each
instance in the image. One point to note about the labels is that
the dimension of the last index, $n_{i}$, depends on the image $i$,
because each image may have a different number of instances.

Our model predicts both a sequence of masks,~$\mathbf{\hat{Y}^{(i)}}=\left\{ \mathbf{\hat{Y}_{1}^{(i)}},\mathbf{\hat{Y}_{2}^{(i)}},\ldots,\mathbf{\hat{Y}_{\hat{n_{i}}}^{(i)}}\right\} $,
for image $i$, where $\mathbf{\hat{Y}_{\hat{t}}^{(i)}}\in[0,1]^{h\times w}$, $\hat{t}\in\{1,\ldots,\hat{n}_{i}\}$,
and a confidence score associated to those masks $\mathbf{s^{(i)}}=\left\{ s_{1}^{(i)},s_{2}^{(i)},\ldots,s_{\hat{n_{i}}}^{(i)}\right\} $.
At inference time, the number of elements predicted, $\hat{n_{i}}$,
depends on the confidence values, $\mathbf{s^{(i)}}$, so that the network
stops producing outputs after time $t$ when $s_{t}^{(i)}<0.5$. At training time we can predefine the length of the
predicted sequence. Given that we know the length, $n_{i}$, of the
$i$-th ground truth annotation, we set the length of the predicted
sequence to be $\hat{n_{i}}=n_{i}+2$, so that the network can learn
when to stop. In any case, the number of elements in the predicted
sequence, $\hat{n}_{i}$, is not necessarily equal to the elements
in the corresponding ground truth set, $n_{i}$, given that our model
could underestimate or overestimate the number of objects.

One way to represent the scenario is to arrange the elements in $\mathbf{Y}$
and $\mathbf{\hat{Y}}$ (where we omit hereafter the index of the image, $i$,
for the sake of clarity) in a bipartite graph, in which each edge
between $\mathbf{Y_{t}}$, $t\in\{1,\ldots n\}$, and $\mathbf{\hat{Y}_{\hat{t}}}$,
$\hat{t}\in\{1,\ldots\hat{n}\}$, has a cost associated to the intersection
over union between $\mathbf{Y_{t}}$ and $\mathbf{\hat{Y}_{\hat{t}}}$. A similarity
measure between $\mathbf{Y}$ and $\mathbf{\hat{Y}}$ can be defined as the maximum
sum of the intersection over union correspondence between the elements
in $\mathbf{Y}$ and the elements in $\mathbf{\hat{Y}}$:

\begin{equation}
\underset{\delta\in\mathcal{S}}{{\rm max}}\,f_{\rm Match}(\mathbf{\hat{Y}},\mathbf{Y},\mathbf{\delta}),
\end{equation}
where
\begin{equation}
f_{\rm Match}(\mathbf{\hat{Y}},\mathbf{Y},\delta)=\overset{\tilde{n}}{\underset{\hat{t}=1}{\sum}}\left(\overset{n}{\underset{t=1}{\sum}}f_{\rm IoU}\left(\mathbf{\hat{Y}_{\hat{t}}},\mathbf{Y_{t}}\right)\delta_{\hat{t},t}\right),\label{eq:Hung}
\end{equation}

\begin{equation}
\mathcal{S}=\left\{ \begin{array}{cc}
\delta\in\{0,1\}^{\tilde{n}\times n}: & \overset{\tilde{n}}{\underset{\hat{t}=1}{\sum}}\delta_{\hat{t},t}\leq1,\,\forall t\in\{1\ldots n\}\\
 & \overset{n}{\underset{t=1}{\sum}}\delta_{\hat{t},t}\leq1,\,\forall\hat{t}\in\{1\ldots\tilde{n}\}
\end{array}\right\} ,\label{eq:Set_S}
\end{equation}
$f_{\rm IoU}(\mathbf{\hat{y}},\mathbf{y})=\frac{\left\langle \mathbf{\hat{y}},\mathbf{y}\right\rangle }{\left\Vert \mathbf{\hat{y}}\right\Vert _{1}+\left\Vert \mathbf{y}\right\Vert _{1}-\left\langle \mathbf{\hat{y}},\mathbf{y}\right\rangle }$,
used in \cite{krahenbuhl2013parameter}, is a relaxed version of the
intersection over union (IoU) that allows the input to take values
in the continuous interval $[0,1]$, and $\tilde{n}={\rm min}(n, \hat{n})$, so that function (\ref{eq:Hung}) only accounts for the first $n$ predicted masks, ignoring the remainder in case our model produces more instances.

The elements in $\delta$ determine the optimal matching between the
elements in $\mathbf{Y}$ and $\mathbf{\hat{Y}}$, so that $\mathbf{\hat{Y}_{\hat{t}}}$ is assigned
to $\mathbf{Y_{t}}$ if and only if $\delta_{\hat{t},t}=1$. The constraint
set $\mathcal{S}$, defined in eq. (\ref{eq:Set_S}), impedes one
ground truth instance being assigned to more than one of the predicted instances
and vice versa. It may be the case that the ground truth $Y_{t}$ is not covered by any prediction
if and only if $\overset{\tilde{n}}{\underset{\hat{t}=1}{\sum}}\delta_{\hat{t},t}=0$.
The optimal matching, $\delta$, can be found out efficiently by means
of the Hungarian algorithm, in a similar vein as in \cite{stewart2015end}.
The coverage loss described in \cite{silberman2014instance} has a similar
form, where the predictions were discrete, and the problem was posed as an
integer program.

\textcolor{black}{End-to-end learning is possible with this loss function, as it is the point-wise minimum of a set of continuous functions (each of those functions corresponding to a possible matching in $\mathcal{S}$). Thus, a direction of decrease of the loss function at a point can be computed by following two steps:
\begin{inparaenum}
\item Figuring out which function in the set $\mathcal{S}$ achieves the minimum at that point.
\item Computing the gradient of that function.
\end{inparaenum}
Here, the Hungarian algorithm is employed in the described first step to find out the function that achieves the minimum at the point. Then, the gradient of that function is computed. The details of this process are shown in Sec. D in the Appendix.}

We now need to account for the confidence scores $\mathbf{s}$ predicted by the model. To do so, we consider
that the ideal output is to predict $s_t=1$ if the number of instances $t$ segmented so far is equal or less than the total number of instances $n$, otherwise $s_t$ should be $0$.
Taking this into account, we propose the following
loss function:
\begin{equation}
\ell(\mathbf{\hat{Y}},\mathbf{s},\mathbf{Y})=  \underset{\delta\in\mathcal{S}}{{\rm min}}-\overset{\tilde{n}}{\underset{\hat{t}=1}{\sum}}\overset{n}{\underset{t=1}{\sum}}f_{\rm IoU}\left(\mathbf{\hat{Y}_{\hat{t}}},\mathbf{Y_{t}}\right)\delta_{\hat{t},t} + \lambda \overset{\hat{n}}{\underset{t=1}{\sum}} f_{\rm BCE}\left([t\leq n],s_t\right),
\label{eq:loss}
\end{equation}
where $f_{\rm BCE}(a,b)=-\left(a{\rm log}(b)+(1-a){\rm log}(1-b)\right)$
is the binary cross entropy, and the Iverson bracket $[\cdot]$ is $1$ if the condition 
within the brackets is true, and $0$ otherwise.
Finally,
$\lambda$ is a hyperparameter that ponders the importance of the
second term with respect to the first one. 

\section{Experiments}

We perform two kinds of experiments, in which we study the capabilities of our approach to both segment and count instances. In the first experiment we focus on multi-instance subject segmentation, and in the second we focus on segmenting and counting leaves in plants. Before presenting those results, we first describe the implementations details that are common to both experiments.

\subsection{Implementation Details of our Method\label{sub:Implementation}}

We have implemented our approach using the Lua/Torch deep learning framework
\cite{collobert2011torch7}. The code and models are publicly available\footnote{Available at \texttt{http://romera-paredes.com/ris}.}.

The recurrent stage is composed of two ConvLSTM layers, so that the output of the first ConvLSTM acts as the input for the second one. This stage is followed by the spatial inhibition module which produces a confidence score together with an instance segmentation mask. The resultant prediction is evaluated according to the loss function defined in eq. (\ref{eq:loss}), where we set $\lambda=1$.

At training stage, the parameters of the recurrent structure are learned by
backpropagation through time. In order
to prevent the exploding gradient effect, we clipped the gradients
so that each of its elements has a maximum absolute value of $5$.
We use the Adam optimization algorithm \cite{kingma2014adam} for
training the whole network,
 setting the initial learning rate to $10^{-4}$, and multiplying it by $0.1$ when the training error plateaus. 
We use neither dropout nor 
$\ell_2$ regularization, as we did not observe overfitting in preliminary experiments. In the same
way as in \cite{Zheng_2015}, we have used one image per batch.

The weights of the recurrent structure are initialized at random,
sampling them uniformly from the interval $[-0.08,0.08]$ with
the exception of the bias terms in the forget gate, $\mathbf{b_f}$ in eq. (\ref{eq:convlstm}). They have been
initialized to $1$ with the aim of allowing by default to backpropagate
the error to previous iterations in the sequence.

We perform curriculum learning by gradually increasing the number of objects that are required to be segmented from the images.
That is, at the beginning we use only $2$ recurrent iterations, so that the network is expected to learn to extract at most $2$
objects per image, even when there are more. Once the training procedure converges, we increment this number, and keep
iterating the process.

At inference time we assign a pixel to an instance if the predicted
value is higher than $0.5$. Nevertheless, we observe that the predicted
pixels values in $\mathbf{\hat{Y}}$ are usually saturated, that is, they are
either very close to $0$ or very close to $1$. Although uncommon, it might happen that
the same pixel is assigned to more than one instance in the sequence.
Whenever that is the case, we assign the pixel to the instance belonging
to the earlier iteration. Finally, the produced sequence terminates whenever the confidence 
score predicted by the network is below $0.5$.

\subsection{Multiple Person Segmentation}
We assess the quality of our approach for detecting and segmenting individual subjects in real images.
Multiple person segmentation is extremely challenging because people in pictures present high variations such as different posture, age, gender, clothing, location and depth within the scene, among others. 

In order to learn this task, we have integrated our model on the FCN-8s network developed in \cite{long2015fully}. The FCN-8s network is composed of a series of layers and adding skips that produce, as a result, an image representation whose size is smaller than the original image. This is followed by an upsampling layer that resizes the representation back to the original image size. We modify this structure by putting the ConvLSTM before this upsampling layer, which is integrated into the subsequent spatial inhibition module, as shown in Fig. \ref{fig:Inhibition}.
Following other works such as \cite{long2015fully,Zheng_2015} we add padding and/or resize the input image so that the resultant size is $500\times 500$. As a consequence, the size of the input of the ConvLSTM layers, as well as their hidden states have dimensionality $64\times 64\times 100$, where $100$ is the number of features extracted for each pixel. All gates ($\mathbf{i_t}$, $\mathbf{f_t}$, $\mathbf{o_t}$, and $\mathbf{g_t}$ in eq. \ref{eq:convlstm}) of both ConvLSTM layers use $1\times 1$ convolutions. 

For training we used the MSCOCO dataset \cite{lin2014microsoft}, and the training images of the Pascal VOC 2012 dataset. We first fixed the weights of the FCN-8s except for the last layer, and then learned the parameters of that last layer, together with the ConvLSTM and the spatial inhibition module, following the procedure described in Sec. \ref{sub:Implementation}. Then we fine-tuned the whole network using a learning rate of $10^{-6}$ until convergence.

We observed that the predictions obtained by recurrent instance segmentation (RIS), while promising, were coarse with respect to
the boundaries of the segmented subjects. That is expected, as the ConvLSTM operates on a low resolution representation of the image. In order to amend this, we have used a CRF as a post-processing method over the produced segments. We call this approach RIS+CRF.

We compare these two approaches with the recent instance segmentation methods presented in \cite{liu2015multi,hariharan2014simultaneous,liang2015proposal}, already introduced in Sec. \ref{sub:instance_seg_review}. 
We also compare to a baseline consisting in performing proposal generation from the semantic segmentation result produced by the FCN-8s of \cite{long2015fully}. We have used Faster R-CNN \cite{ren2015faster} as the proposal generation method.
Following previous works, we measure the predictive performance of the methods with respect to the Pascal VOC 2012 validation set, using two standard metrics: average precision ($AP^r$) on the predicted regions having over $0.5$ IoU overlapping with ground truth masks, denoted as $AP^r (0.5)$; and averaging the $AP^r$ for different degrees of IoU overlapping, from $0.1$ to $0.9$, denoted as $AP^r Ave$.
We show the results in Table \ref{tab:results}. We observe that RIS achieves comparable results to state of the art approaches. When using CRF as a post-processing method the results improve, outperforming the competing methods. We also provide some qualitative results in Fig. \ref{fig:future}, and more extensively in Sec. C in the Appendix.

\begin{table}
\begin{centering}
\begin{tabular}{|c|c|c|c|c|c|c|}
\hline 
& Baseline & \cite{liu2015multi} & \cite{hariharan2014simultaneous} & \cite{liang2015proposal} & RIS & RIS+CRF \tabularnewline
\hline 
\hline 
$AP^r (0.5)$ & $45.8$ & $48.3$ & $47.9$ & $48.8$ & $46.7$ & $\mathbf{50.1}$ \tabularnewline
\hline 
\hline 
$AP^r Ave$ & $39.6$ & $$ & $$ & $42.9$ & $41.9$ & $\mathbf{43.7}$ \tabularnewline
\hline 
\end{tabular}
\par\end{centering}
\caption{\label{tab:results} Multiple person segmentation comparison with state of the art approaches on the PASCAL VOC 2012 validation set. First row: using $AP^r$ metric at $0.5$ IoU. Second row: averaging $AP^r$ metric from 0.1 to 0.9 IoU (gaps indicate unreported results).}
\end{table}

\begin{figure}
\begin{centering}
\begin{tabular}{ccc}
\includegraphics[width=0.25\columnwidth]{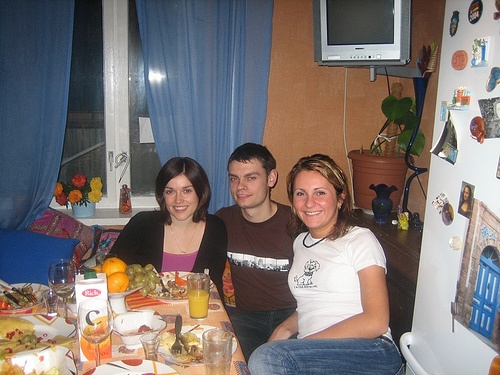} & \includegraphics[width=0.25\columnwidth]{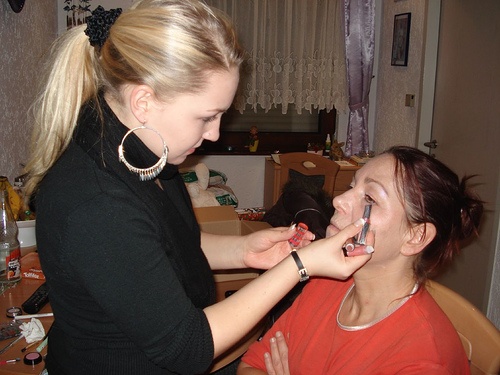} & \includegraphics[width=0.25\columnwidth]{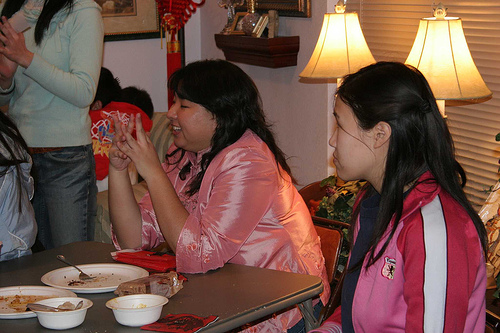}
\tabularnewline
\includegraphics[width=0.25\columnwidth]{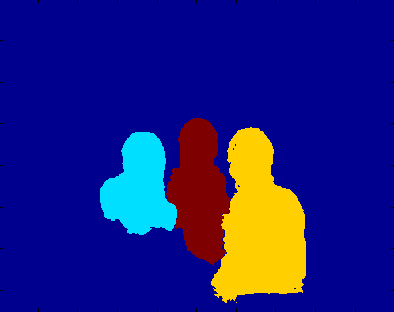} & \includegraphics[width=0.25\columnwidth]{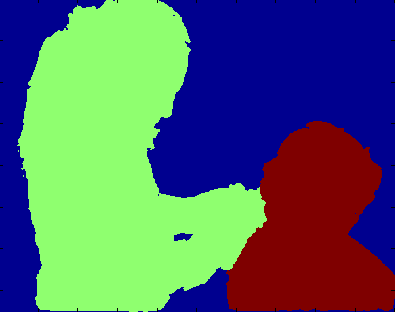} & \includegraphics[width=0.25\columnwidth]{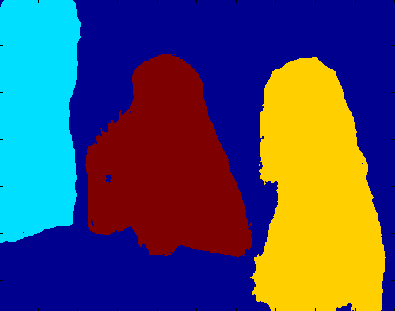}\tabularnewline
\end{tabular}
\par\end{centering}

\caption{\label{fig:future} Instance segmentation for detecting people using
RIS+CRF. Input images taken from the VOC Pascal 2012 dataset. Best viewed in colour.}
\end{figure}

\subsection{Plants Leaf Segmentation and Counting}

Automatic leaf segmentation and counting are useful tasks for plant
phenotyping applications that can lead to improvements in seed production
and plant breeders processes. In this section we use the Computer
Vision Problems in Plant Phenotyping (CVPPP) dataset \cite{minervini2014image,Minervini_PRL2015}.
In particular, we utilize the A1 subset of plants, which is the
biggest subset available, and contains $161$ top-down view images,
having $500\times530$ size each, as the one shown in Fig. (\ref{fig:RIS}, Left).
The training set is composed of $128$ of those images, which have
annotations available. The remaining $33$ images are left out for
testing purposes. All these images are challenging because they present
a high range of variations, with occasional leaf occlusions, varied
backgrounds, several slightly blurred images due to a lack of focus,
and complex leaf shapes.

The limited number of training images has driven us to augment the data
by considering some valid transformations of the images. We apply
two transformations: rotating the image by a random angle, and flipping
the resultant image with a probability of $0.5$.

We learn the fully convolutional network from scratch. Its structure is composed by
a sequence of $5$ convolutional layers, each of them followed by a rectified linear unit. The first convolution
learns $30$ $9\times 9$ filters, and the following four learn $30$ $3\times 3$ filters each.
This sequence of convolutions produces a $100\times106\times30$ representation of the image, which is the input to the recurrent stage of the model. In the stack of two ConvLSTMs we have set all gates to have $3\times 3$ convolutional layers. 


We compare our method with several approaches submitted to the CVPPP
challenges on both leaf segmentation and counting. These are:
\begin{itemize}
\item IPK Gatersleben \cite{pape20143}: Firstly, it segments foreground
from background by using 3D histograms and a supervised learning model.
Secondly, it identifies the leaves centre points and leaves split
points by applying unsupervised learning methods, and then it segments
individual leaves by applying graph-based noise removal, and region
growing techniques.
\item Nottingham \cite{Collation}: Firstly, SLIC is applied on the image
in order to get superpixels. Then the plant is extracted from the
background. The superpixels in the centroids of each leaf are identified
by finding local maxima on the distance map of the foreground. Finally,
leaves are segmented by applying the watershed transform.
\item MSU \cite{Collation}: It is based on aligning instances (leaves)
with a given set of templates by using Chamfer Matching \cite{barrow1977parametric}.
In this case the templates are obtained from the training set ground
truth.
\item Wageningen \cite{yin2014multi}: It performs foreground segmentation
using a neural network, followed by a series of image processing transformations,
including inverse distance image transform from the detected foreground,
and using the watershed transform to segment leaves individually.
\item PRIAn \cite{giuffrida2015learning}: First, a set of features is learned
in an unsupervised way on a log-polar representation of the image.
Then, a support vector regression model is applied on the resultant
features in order to predict the number of leaves.
\end{itemize}
These competing methods are explicitly designed to perform well in
this particular plant leaf segmentation and counting problems, containing heuristics
that are only valid on this domain. On the contrary, the
applicability of our model is broad.
\begin{table}
\noindent \begin{centering}
\begin{tabular}{|c|c|c|c|c|c|c|c|}
\hline 
 & IPK & Nottingham & MSU & Wageningen & PRIAn & RIS & RIS+CRF \tabularnewline
\hline 
\hline 
$DiC$ $\longrightarrow0$ & $\begin{array}{c}
-1.9\\
(2.5)
\end{array}$ & $\begin{array}{c}
-3.6\\
(2.4)
\end{array}$ & $\begin{array}{c}
-2.3\\
(1.6)
\end{array}$ & $\begin{array}{c}
-0.4\\
(3.0)

\end{array}$ & $\begin{array}{c}
0.8\\
(1.5)
\end{array}$ & $\begin{array}{c}
\boldsymbol{0.2}\\
\boldsymbol{\boldsymbol{(1.4)}}
\end{array}$ & $\begin{array}{c}
\boldsymbol{0.2}\\
\boldsymbol{\boldsymbol{(1.4)}}
\end{array}$ \tabularnewline
\hline 
$\left|DiC\right|$$\downarrow$ & $\begin{array}{c}
2.6\\
(1.8)
\end{array}$ & $\begin{array}{c}
3.8\\
(2.0)
\end{array}$ & $\begin{array}{c}
2.3\\
(1.5)
\end{array}$ & $\begin{array}{c}
2.2\\
(2.0)
\end{array}$ & $\begin{array}{c}
1.3\\
(2.0)
\end{array}$ & $\begin{array}{c}
\boldsymbol{1.1}\\
\boldsymbol{(0.9)}
\end{array}$ & $\begin{array}{c}
\boldsymbol{1.1}\\
\boldsymbol{(0.9)}
\end{array}$\tabularnewline
\hline 
$SBD(\%)\uparrow$ & $\begin{array}{c}
\boldsymbol{74.4}\\
\boldsymbol{(4.3)}
\end{array}$ & $\begin{array}{c}
68.3\\
(6.3)
\end{array}$ & $\begin{array}{c}
66.7\\
(7.6)
\end{array}$ & $\begin{array}{c}
71.1\\
(6.2)
\end{array}$ & $\begin{array}{c}
-
\end{array}$ & $\begin{array}{c}
56.8\\
(8.2)
\end{array}$ & $\begin{array}{c}
66.6\\
(8.7)
\end{array}$\tabularnewline
\hline 
\end{tabular}
\par\end{centering}

\caption{\label{tab:plant_results} Results obtained on the CVPPP dataset according to the measures: Difference in Count ($DiC$), absolute Difference in Count ($\left|DiC\right|$), and Symmetric Best Dice ($SBD$). Reported mean and standard deviation (in parenthesis).}
\end{table}

The results obtained are shown in Table \ref{tab:plant_results},
using the measures reported by the CVPPP organization in order to
compare the submitted solutions. These are 
Difference in Count
($DiC$) which is the difference between the predicted number of leaves
and the ground truth, $\left|DiC\right|$, which is the absolute
value of DiC averaged across all images, and Symmetric Best Dice ($SBD$),
which is defined in \cite{Collation}, and provides a measure about
the accuracy of the segmentation of the instances.
Regarding leaf counting, we observe that our approach
significantly outperforms the competing ad hoc methods that were designed for this particular problem.
With regard to segmentation of leaves, our approach obtains comparable results with
respect to the competitors, yet it does not outperform any approach.
We hypothesize that despite the data augmentation process we follow, the amount
of original images available for training is too small as to learn how to segment a wide variety of leaf shapes from scratch. This scarcity of training data has a smaller impact in the competing methods, given that they contain heuristics and prior information about the problem. 

\begin{figure}
\begin{centering}
\begin{tabular}{cccccc}
$t$ & $1$ & $2$ & $3$ & $4$ & $5$\tabularnewline
\hline 
$\kappa(\mathbf{h_t})$ & \includegraphics[width=0.12\columnwidth]{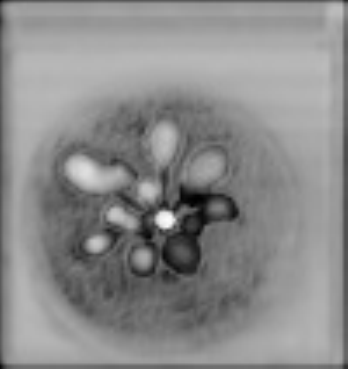} & \includegraphics[width=0.12\columnwidth]{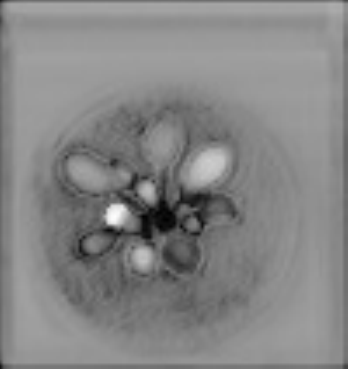} & \includegraphics[width=0.12\columnwidth]{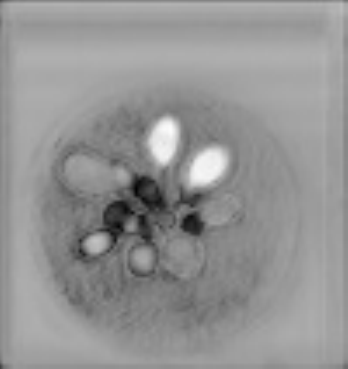} & \includegraphics[width=0.12\columnwidth]{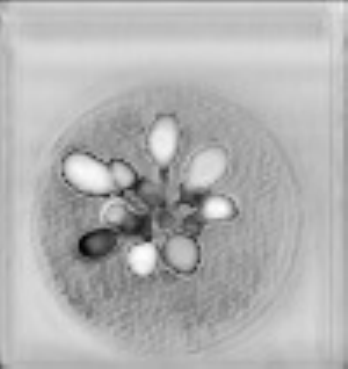} & \includegraphics[width=0.12\columnwidth]{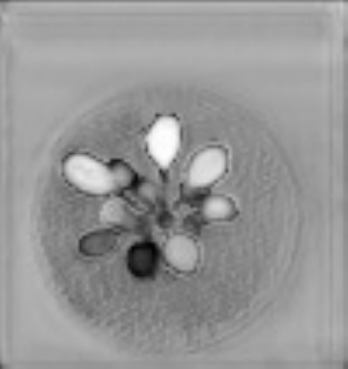}\tabularnewline
$\mathbf{\hat{Y}_t}$ & \includegraphics[width=0.12\columnwidth]{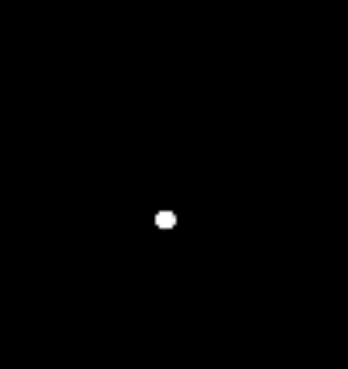} & \includegraphics[width=0.12\columnwidth]{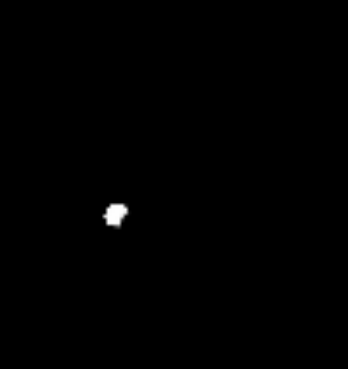} & \includegraphics[width=0.12\columnwidth]{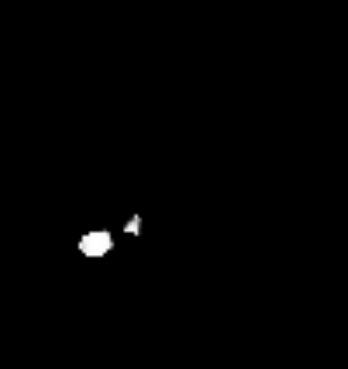} & \includegraphics[width=0.12\columnwidth]{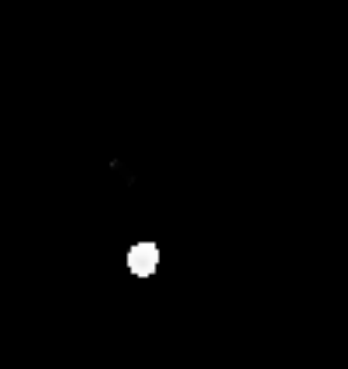} & \includegraphics[width=0.12\columnwidth]{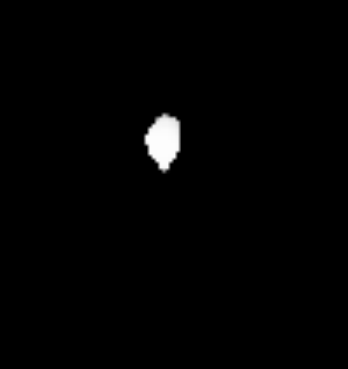}\tabularnewline
\end{tabular}
\par\end{centering}

\caption{\label{fig:plants_exp} Representation of the state, $\kappa(\mathbf{h_t})$ (the sum of the absolute values across channels), and the output, $\mathbf{\hat{Y}_t}$, of a sequence produced by our model. 
}
\end{figure}

We have also visualized, in Fig. \ref{fig:plants_exp}, a representation
of what the network keeps in memory as the sequence is produced. The
column denoted as $\kappa(\mathbf{h_t})$ shows a summary function of the hidden state
$\mathbf{h}$ of the second ConvLSTM layer. The summary function consists of the sum of the absolute values across channels for each pixel.
The column denoted as $\mathbf{\hat{Y}_t}$ corresponds to the output produced
by the network. We observe that as time advances, the state is modified
to take into account parts of the image that have been visited. We also inspected the value of the cell, that is $\mathbf{c_t}$ in eq. (\ref{eq:convlstm}), but we have not observed any clear clues. 

\section{Discussion}

In this paper we have proposed a new instance segmentation paradigm
characterized by its sequential nature. Similarly to what human beings
do when counting objects in a scene, our model proceeds sequentially,
segmenting one instance of the scene at a time. The resulting model
integrates in a single pipeline all the required functions to segment
instances. These functions are defined by a set of parameters that
are jointly learned end-to-end. A key aspect in our model is the use
of a recurrent structure that is able to track visited areas in the
image as well as to handle occlusion among instances. Another key
aspect is the definition of a loss function that accurately represents
the instance segmentation objective we aim to achieve. The experiments
carried out on multiple person segmentation and leaf counting show that our approach outperforms state of the art methods. Qualitative results show that the state
in the recurrent stage contains information regarding the visited
instances in the sequence.

The primary objective of this paper is to show that learning end-to-end
instance segmentation is possible by means of a recurrent neural network. Nevertheless, variations of some architectural choices could lead to
even better results. We tried a variety of other alternatives, such as adding
the sum of the prediction masks as an extra input into the recurrent unit. We
also tried alternatives to $f_{\rm IoU}$ in eq. (\ref{eq:Hung}), such as log-likelihood. The results obtained in either case were not better than
the ones achieved by the model described in Sec. \ref{sec:model}. The analysis
of these and other alternative architectures is left for future work.

There are several extensions that can be carried out in our approach.
One is allowing the model to classify the segmented instance at each
time. This can be
done by generalizing the loss function in eq. (\ref{eq:loss}). Another extension consists in integrating the CRF module as a layer 
in the end-to-end model, such as in \cite{Zheng_2015}.
Another interesting line of research is investigating other recurrent structures that could be as good or even better than ConvLSTMs for instance segmentation.
Finally, the extension of this model for exploiting co-occurrence of objects, parts of objects, and attributes, could be a promising research direction.

\section*{Acknowledgments}
This work was supported by the EPSRC, ERC
grant ERC-2012-AdG 321162-HELIOS, HELIOS-DFR00200, EPSRC grant Seebibyte EP/M013774/1 and
EPSRC/MURI grant EP/N019474/1.

Thanks to Siddharth Narayanaswamy for proofreading this work.
\clearpage

\bibliographystyle{splncs}
\bibliography{RIS}

\begin{thebibliography}{10}

\bibitem{smart}
VA-ST:
\newblock Smart specs.
\newblock \url{http://www.va-st.com/smart-specs}

\bibitem{arteta2013learning}
Arteta, C., Lempitsky, V., Noble, J.A., Zisserman, A.:
\newblock Learning to detect partially overlapping instances.
\newblock In: Computer Vision and Pattern Recognition (CVPR), IEEE (2013)
  3230--3237

\bibitem{trager1976human}
Trager, W., Jensen, J.B.:
\newblock Human malaria parasites in continuous culture.
\newblock Science \textbf{193}(4254) (1976)  673--675

\bibitem{hariharan2014simultaneous}
Hariharan, B., Arbel{\'a}ez, P., Girshick, R., Malik, J.:
\newblock Simultaneous detection and segmentation.
\newblock In: European Conference on Computer Vision (ECCV).
\newblock Springer (2014)  297--312

\bibitem{liu2015multi}
Chen, Y.T., Liu, X., Yang, M.H.:
\newblock Multi-instance object segmentation with occlusion handling.
\newblock In: Proceedings of the IEEE Conference on Computer Vision and Pattern
  Recognition (CVPR). (2015)  3470--3478

\bibitem{vineet2011human}
Vineet, V., Warrell, J., Ladicky, L., Torr, P.H.:
\newblock Human instance segmentation from video using detector-based
  conditional random fields.
\newblock In: British Machine Vision Conference (BMVC). (2011)  1--11

\bibitem{dehaene1994dissociable}
Dehaene, S., Cohen, L.:
\newblock Dissociable mechanisms of subitizing and counting: neuropsychological
  evidence from simultanagnosic patients.
\newblock Journal of Experimental Psychology: Human Perception and Performance
  \textbf{20}(5) (1994)  958

\bibitem{porter2007effort}
Porter, G., Troscianko, T., Gilchrist, I.D.:
\newblock Effort during visual search and counting: Insights from pupillometry.
\newblock The Quarterly Journal of Experimental Psychology \textbf{60}(2)
  (2007)  211--229

\bibitem{ladicky2010and}
Ladick{\`y}, L., Sturgess, P., Alahari, K., Russell, C., Torr, P.H.:
\newblock What, where and how many? combining object detectors and crfs.
\newblock In: European Conference on Computer Vision (ECCV).
\newblock Springer (2010)  424--437

\bibitem{girshick2014rich}
Girshick, R., Donahue, J., Darrell, T., Malik, J.:
\newblock Rich feature hierarchies for accurate object detection and semantic
  segmentation.
\newblock In: Computer Vision and Pattern Recognition (CVPR), IEEE (2014)
  580--587

\bibitem{arbelaez2014multiscale}
Arbelaez, P., Pont-Tuset, J., Barron, J., Marques, F., Malik, J.:
\newblock Multiscale combinatorial grouping.
\newblock In: Computer Vision and Pattern Recognition (CVPR), 2014 IEEE
  Conference on, IEEE (2014)  328--335

\bibitem{long2015fully}
Long, J., Shelhamer, E., Darrell, T.:
\newblock Fully convolutional networks for semantic segmentation.
\newblock In: Proceedings of the IEEE Conference on Computer Vision and Pattern
  Recognition (CVPR). (2015)  3431--3440

\bibitem{chen2014semantic}
Liang-Chieh, C., Papandreou, G., Kokkinos, I., Murphy, K., Yuille, A.:
\newblock Semantic image segmentation with deep convolutional nets and fully
  connected crfs.
\newblock In: International Conference on Learning Representations (ICLR).
  (2015)

\bibitem{Zheng_2015}
Zheng, S., Jayasumana, S., Romera-Paredes, B., Vineet, V., Su, Z., Du, D.,
  Huang, C., Torr, P.H.:
\newblock Conditional random fields as recurrent neural networks.
\newblock IEEE International Conference on Computer Vision (ICCV) (2015)

\bibitem{liang2015proposal}
Liang, X., Wei, Y., Shen, X., Yang, J., Lin, L., Yan, S.:
\newblock Proposal-free network for instance-level object segmentation.
\newblock arXiv preprint arXiv:1509.02636 (2015)

\bibitem{tighe2014scene}
Tighe, J., Niethammer, M., Lazebnik, S.:
\newblock Scene parsing with object instances and occlusion ordering.
\newblock In: Computer Vision and Pattern Recognition (CVPR), IEEE (2014)
  3748--3755

\bibitem{yang2012layered}
Yang, Y., Hallman, S., Ramanan, D., Fowlkes, C.C.:
\newblock Layered object models for image segmentation.
\newblock Pattern Analysis and Machine Intelligence, IEEE Transactions on
  \textbf{34}(9) (2012)  1731--1743

\bibitem{zhang2015monocular}
Zhang, Z., Schwing, A.G., Fidler, S., Urtasun, R.:
\newblock Monocular object instance segmentation and depth ordering with cnns.
\newblock In: IEEE International Conference on Computer Vision (ICCV).
\newblock (2015)  2614--2622

\bibitem{bahdanau2014neural}
Bahdanau, D., Cho, K., Bengio, Y.:
\newblock Neural machine translation by jointly learning to align and
  translate.
\newblock arXiv preprint arXiv:1409.0473 (2014)

\bibitem{graves2009offline}
Graves, A., Schmidhuber, J.:
\newblock Offline handwriting recognition with multidimensional recurrent
  neural networks.
\newblock In: Advances in Neural Information Processing Systems (NIPS). (2009)
  545--552

\bibitem{vinyals2015neural}
Vinyals, O., Le, Q.:
\newblock A neural conversational model.
\newblock arXiv preprint arXiv:1506.05869 (2015)

\bibitem{gregor2015draw}
Gregor, K., Danihelka, I., Graves, A., Wierstra, D.:
\newblock Draw: A recurrent neural network for image generation.
\newblock Proceedings of the 32nd International Conference on Machine Learning
  (ICML) (2015)

\bibitem{donahue2015long}
Donahue, J., Anne~Hendricks, L., Guadarrama, S., Rohrbach, M., Venugopalan, S.,
  Saenko, K., Darrell, T.:
\newblock Long-term recurrent convolutional networks for visual recognition and
  description.
\newblock In: Proceedings of the IEEE Conference on Computer Vision and Pattern
  Recognition (CVPR). (2015)  2625--2634

\bibitem{karpathy2014deep}
Karpathy, A., Fei-Fei, L.:
\newblock Deep visual-semantic alignments for generating image descriptions.
\newblock In: Proceedings of the IEEE Conference on Computer Vision and Pattern
  Recognition (CVPR). (2015)  3128--3137

\bibitem{vinyals2015show}
Vinyals, O., Toshev, A., Bengio, S., Erhan, D.:
\newblock Show and tell: A neural image caption generator.
\newblock In: Conference on Computer Vision and Pattern Recognition (CVPR).
  (2015)  3156--3164

\bibitem{winther2015convolutional}
Winther, O.:
\newblock Convolutional lstm networks for subcellular localization of proteins.
\newblock In: Algorithms for Computational Biology: Second International
  Conference, AlCoB 2015, Mexico City, Mexico, August 4-5, 2015, Proceedings.
  Volume 9199., Springer (2015) ~68

\bibitem{stewart2015end}
Stewart, R., Andriluka, M.:
\newblock End-to-end people detection in crowded scenes.
\newblock arXiv preprint arXiv:1506.04878 (2015)

\bibitem{pavel2015recurrent}
Pavel, M.S., Schulz, H., Behnke, S.:
\newblock Recurrent convolutional neural networks for object-class segmentation
  of rgb-d video.
\newblock In: International Joint Conference on Neural Networks (IJCNN), IEEE
  (2015)  1--8

\bibitem{williams2004greedy}
Williams, C.K., Titsias, M.K.:
\newblock Greedy learning of multiple objects in images using robust statistics
  and factorial learning.
\newblock Neural Computation \textbf{16}(5) (2004)  1039--1062

\bibitem{ba2014multiple}
Ba, J., Mnih, V., Kavukcuoglu, K.:
\newblock Multiple object recognition with visual attention.
\newblock International Conference on Learning Representations (ICLR) (2015)

\bibitem{xu2015show}
Xu, K., Ba, J., Kiros, R., Cho, K., Courville, A., Salakhudinov, R., Zemel, R.,
  Bengio, Y.:
\newblock Show, attend and tell: Neural image caption generation with visual
  attention.
\newblock In: Proceedings of the 32nd International Conference on Machine
  Learning (ICML).
\newblock (2015)  2048--2057

\bibitem{mnih2014recurrent}
Mnih, V., Heess, N., Graves, A.,  et~al.:
\newblock Recurrent models of visual attention.
\newblock In: Advances in Neural Information Processing Systems (NIPS). (2014)
  2204--2212

\bibitem{hermann2015teaching}
Hermann, K.M., Kocisky, T., Grefenstette, E., Espeholt, L., Kay, W., Suleyman,
  M., Blunsom, P.:
\newblock Teaching machines to read and comprehend.
\newblock In: Advances in Neural Information Processing Systems (NIPS). (2015)
  1693--1701

\bibitem{hochreiter1997long}
Hochreiter, S., Schmidhuber, J.:
\newblock Long short-term memory.
\newblock Neural computation \textbf{9}(8) (1997)  1735--1780

\bibitem{fischer2015flownet}
Dosovitskiy, A., Fischery, P., Ilg, E., Hazirbas, C., Golkov, V., van~der
  Smagt, P., Cremers, D., Brox, T.,  et~al.:
\newblock Flownet: Learning optical flow with convolutional networks.
\newblock In: IEEE International Conference on Computer Vision (ICCV), IEEE
  (2015)  2758--2766

\bibitem{shi2015convolutional}
Xingjian, S., Chen, Z., Wang, H., Yeung, D.Y., Wong, W.k., Woo, W.c.:
\newblock Convolutional lstm network: A machine learning approach for
  precipitation nowcasting.
\newblock In: Advances in Neural Information Processing Systems (NIPS).
\newblock (2015)  802--810

\bibitem{krahenbuhl2013parameter}
Kr{\"a}henb{\"u}hl, P., Koltun, V.:
\newblock Parameter learning and convergent inference for dense random fields.
\newblock In: Proceedings of the 30th International Conference on Machine
  Learning (ICML). (2013)  513--521

\bibitem{silberman2014instance}
Silberman, N., Sontag, D., Fergus, R.:
\newblock Instance segmentation of indoor scenes using a coverage loss.
\newblock In: European Conference on Computer Vision (ECCV).
\newblock Springer (2014)  616--631

\bibitem{collobert2011torch7}
Collobert, R., Kavukcuoglu, K., Farabet, C.:
\newblock Torch7: A matlab-like environment for machine learning.
\newblock In: BigLearn, NIPS Workshop. Number EPFL-CONF-192376 (2011)

\bibitem{kingma2014adam}
Kingma, D., Ba, J.:
\newblock Adam: A method for stochastic optimization.
\newblock arXiv preprint arXiv:1412.6980 (2014)

\bibitem{lin2014microsoft}
Lin, T.Y., Maire, M., Belongie, S., Hays, J., Perona, P., Ramanan, D.,
  Doll{\'a}r, P., Zitnick, C.L.:
\newblock Microsoft coco: Common objects in context.
\newblock In: European Conference on Computer Vision (ECCV).
\newblock Springer (2014)  740--755

\bibitem{ren2015faster}
Ren, S., He, K., Girshick, R., Sun, J.:
\newblock Faster r-cnn: Towards real-time object detection with region proposal
  networks.
\newblock In: Advances in neural information processing systems (NIPS). (2015)
  91--99

\bibitem{minervini2014image}
Minervini, M., Abdelsamea, M.M., Tsaftaris, S.A.:
\newblock Image-based plant phenotyping with incremental learning and active
  contours.
\newblock Ecological Informatics \textbf{23} (2014)  35--48

\bibitem{Minervini_PRL2015}
Minervini, M., Fischbach, A., Scharr, H., Tsaftaris, S.A.:
\newblock Finely-grained annotated datasets for image-based plant phenotyping.
\newblock Pattern Recognition Letters (2015) Special Issue on Fine-grained
  Categorization in Ecological Multimedia.

\bibitem{pape20143}
Pape, J.M., Klukas, C.:
\newblock 3-d histogram-based segmentation and leaf detection for rosette
  plants.
\newblock In: Computer Vision-ECCV 2014 Workshops, Springer (2014)  61--74

\bibitem{Collation}
Scharr, H., Minervini, M., French, A.P., Klukas, C., Kramer, D.M., Liu, X.,
  Luengo~Muntion, I., Pape, J.M., Polder, G., Vukadinovic, D., Yin, X.,
  Tsaftaris, S.A.:
\newblock Leaf segmentation in plant phenotyping: A collation study.
\newblock Machine Vision and Applications (In print) (2015)

\bibitem{barrow1977parametric}
Barrow, H.G., Tenenbaum, J.M., Bolles, R.C., Wolf, H.C.:
\newblock Parametric correspondence and chamfer matching: Two new techniques
  for image matching.
\newblock Technical report, DTIC Document (1977)

\bibitem{yin2014multi}
Yin, X., Liu, X., Chen, J., Kramer, D.M.:
\newblock Multi-leaf tracking from fluorescence plant videos.
\newblock In: Image Processing (ICIP), 2014 IEEE International Conference on,
  IEEE (2014)  408--412

\bibitem{giuffrida2015learning}
Giuffrida, M.V., Minervini, M., Tsaftaris, S.A.:
\newblock Learning to count leaves in rosette plants, British Machine Vision
  Conference (CVPPP Workshop). BMVA Press (2015)

\end{thebibliography}

\clearpage
\appendix
\section*{Appendix}

Here we extend the content of the paper in four ways. Firstly, we visualize the feature representations and RNN states of the model, when it processes a given image. Secondly, we depict the update equations of the ConvLSTM with a diagram. Thirdly we provide many examples, including failure cases, of how our model performs on multiple person segmentation. Finally, we provide a more detailed explanation of the loss function introduced in Sec. \ref{sub:Loss-Function}.

\section{Features and States through the Pipeline \label{sec:app_features}}
\begin{figure}[h]
\begin{centering}
\includegraphics[width=0.35\columnwidth]{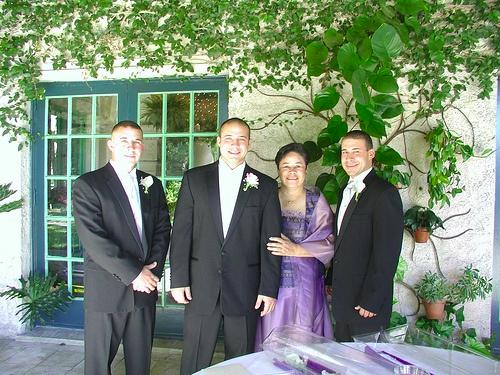} 
\includegraphics[width=0.35\columnwidth]{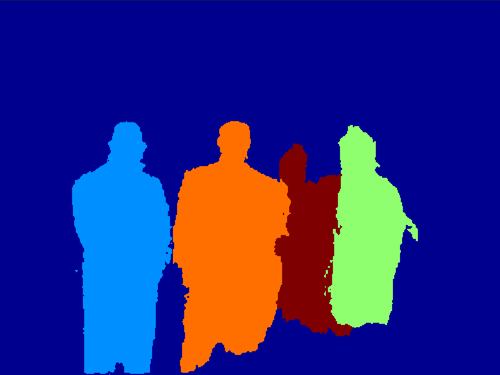} 
\par\end{centering}
\caption{\label{fig:input} Image utilized as an input (Left), and the prediction produced by our model (Right). 
}
\end{figure}

Here we visualize the feature representations built by our model at middle stages of its pipeline. To do so, we utilize the image in Fig. \ref{fig:input} (Left), which is one of the images from the validation Pascal VOC dataset that contained $4$ subjects.

\begin{figure}[h]
\begin{centering}
\includegraphics[width=0.65\columnwidth]{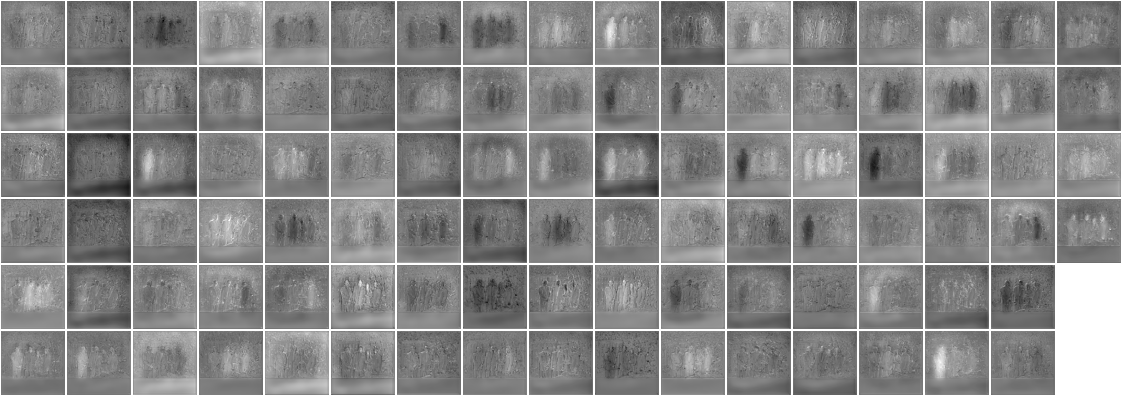} 
\par\end{centering}
\caption{\label{fig:features_people} Features produced by the FCN stage of our model when using the image in Fig. \ref{fig:input} (Left) as an input. 
}
\end{figure}

In Fig. \ref{fig:features_people} we visualize the features extracted in the $100$ output channels of the FCN stage of our model. Here we can see that different instances have different features at this level. Furthermore, some of the channels are somewhat sensitive to some body parts, particularly faces.

\begin{figure}[h]
\begin{centering}
\begin{tabular}{cccccc}
$t$ & $1$ & $2$ & $3$ & $4$ & $5$\tabularnewline
\hline 
$\kappa(\mathbf{h_t})$ & \includegraphics[width=0.12\columnwidth]{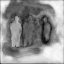} & \includegraphics[width=0.12\columnwidth]{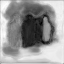} & \includegraphics[width=0.12\columnwidth]{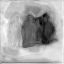} & \includegraphics[width=0.12\columnwidth]{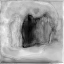} & \includegraphics[width=0.12\columnwidth]{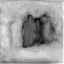}\tabularnewline
${\rm Conv}(\mathbf{h_t})$ & \includegraphics[width=0.12\columnwidth]{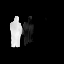} & \includegraphics[width=0.12\columnwidth]{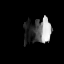} & \includegraphics[width=0.12\columnwidth]{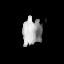} & \includegraphics[width=0.12\columnwidth]{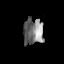} & \includegraphics[width=0.12\columnwidth]{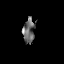}\tabularnewline
$f_{\rm LSM}({\rm Conv}(\mathbf{h_t}))+b$ & \includegraphics[width=0.12\columnwidth]{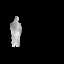} & \includegraphics[width=0.12\columnwidth]{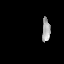} & \includegraphics[width=0.12\columnwidth]{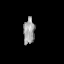} & \includegraphics[width=0.12\columnwidth]{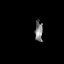} & \includegraphics[width=0.12\columnwidth]{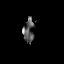}\tabularnewline
$\mathbf{\hat{Y}_t}$ & \includegraphics[width=0.12\columnwidth]{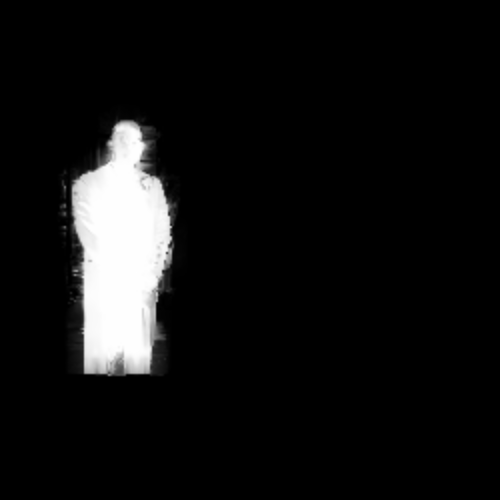} & \includegraphics[width=0.12\columnwidth]{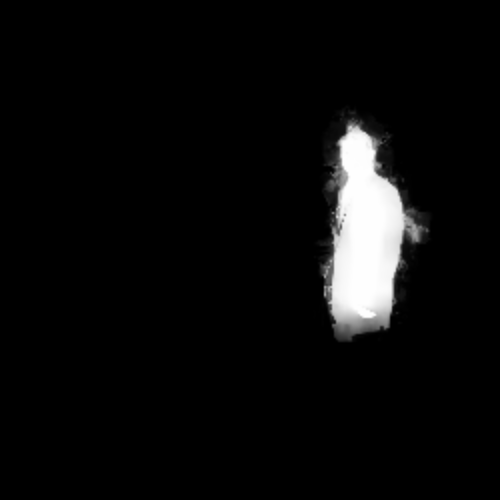} & \includegraphics[width=0.12\columnwidth]{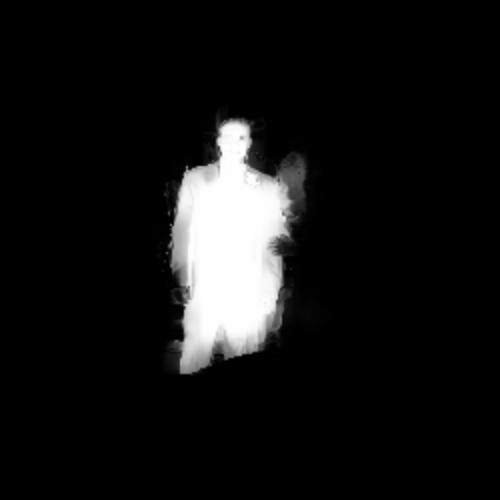} & \includegraphics[width=0.12\columnwidth]{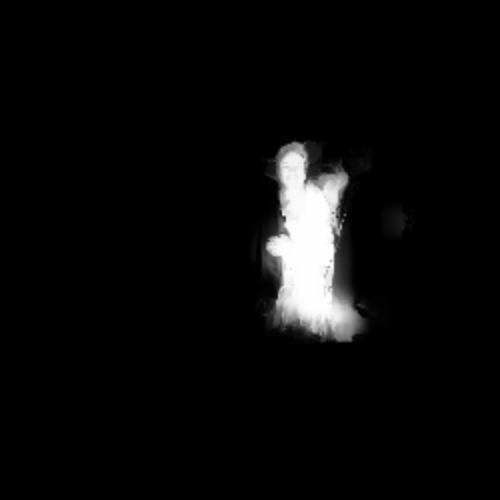} & \includegraphics[width=0.12\columnwidth]{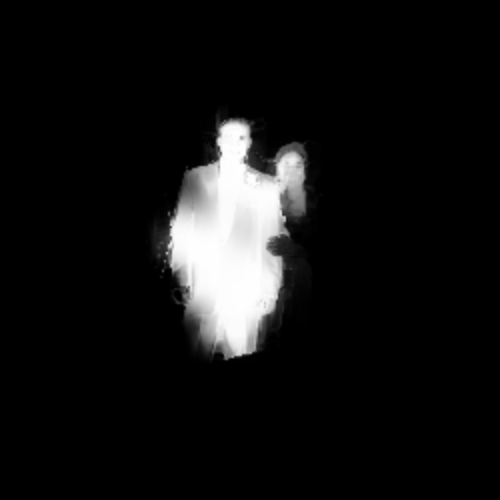}\tabularnewline
$s_t$ & $0.9944$ & $1.0000$ & $0.9998$ & $0.9610$ & $0.4217$\tabularnewline
\end{tabular}
\par\end{centering}
\caption{\label{fig:people_exp} Intermediate representations of the model pipeline when producing a sequence using the image in Fig. \ref{fig:input} (Left) as an input. First row: $\kappa(\mathbf{h_t})$ (the sum of the absolute values across channels). Second and third row: representation at two different stages in the spatial inhibition module. Fourth and fifth: the output $\mathbf{\hat{Y}_t}$ and $s_t$. 
}
\end{figure}

In Fig. \ref{fig:people_exp} we see the state of the ConvLSTM, as well as the representation at two middle stages of the spatial inhibition module, as the model keeps producing segmentations of new instances. We also show the prediction and the confidence score.

In particular, the first row of Fig. \ref{fig:people_exp} shows the absolute values across channels of the state $\mathbf{h_t}$. In the second row, we show the representation obtained after the $1\times 1$ convolution in the spatial inhibition module, which produces a 1-channel output. For the sake of clarity in the visualization, we have filtered out the pixels that have a negative value. We see that at this stage the representation is able to separate one object of interest, but it often contains other elements with lower intensity. In the third row, we show the representation obtained by the spatial inhibition module, just before the sigmoid transformation. Here we have also omitted the negative pixel values. We observe that the representation has successfully filtered out all pixels but the ones belonging to the object of interest. We also observed that the intensity values of the produced mask are not uniform. Finally, the following row shows the masks produced at the end of the pipeline (together with the confidence scores). The sigmoid operation allows the spatial inhibition module to produce pixel values that are saturated, that is, that are close to $1$ if they belong to the object of interest, or close to $0$ if they do not.

\clearpage

\section{Diagram of ConvLSTM \label{sec:app_convlstm}}

Here we provide a diagram illustrating the recurrent update equations of the ConvLSTM (eq. \ref{eq:convlstm}).

\begin{figure*}
\noindent \begin{centering}
\includegraphics[width=0.6\paperwidth]{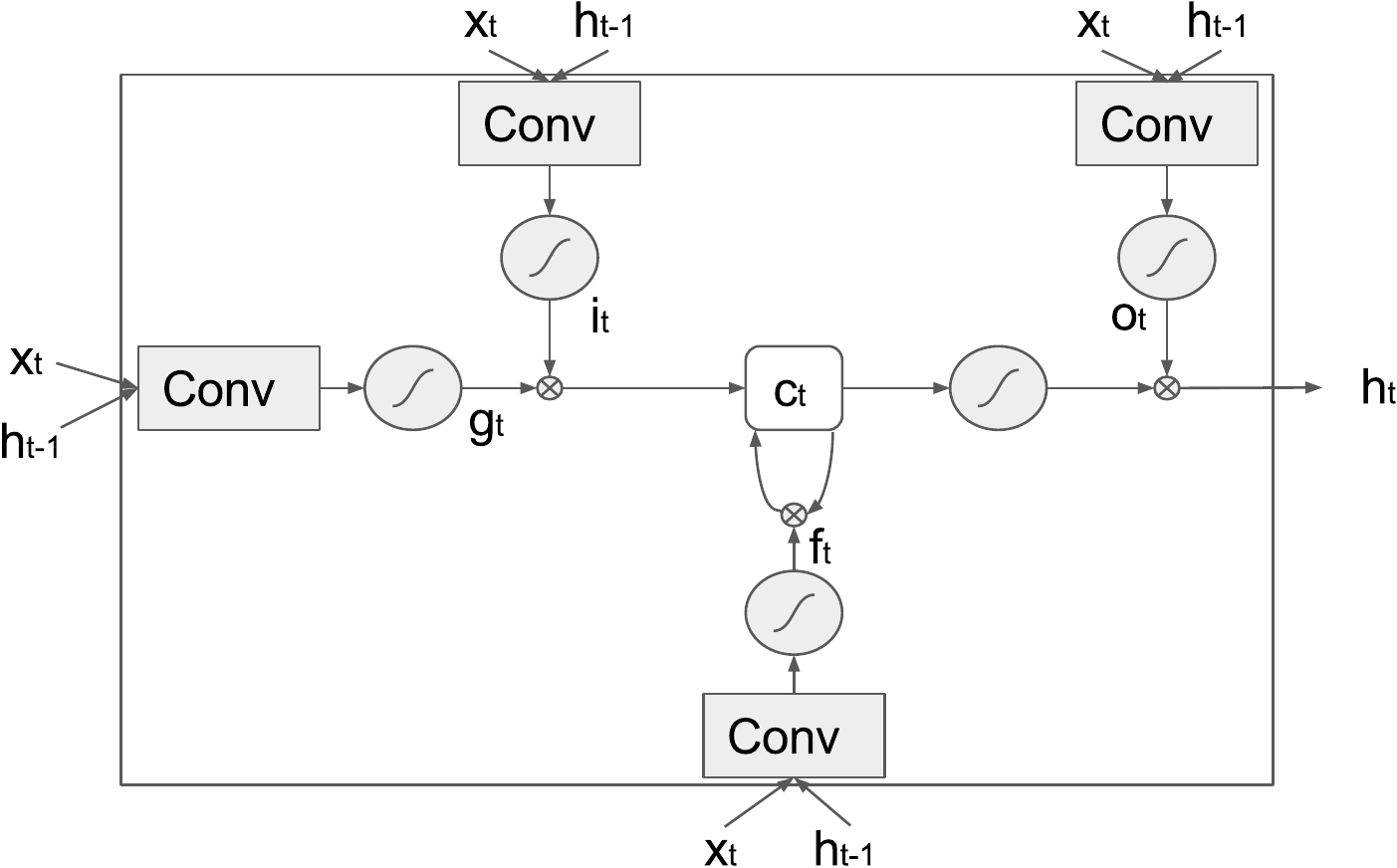}
\par\end{centering}

\caption{\label{fig:ConvLSTM} Diagram of a ConvLSTM.}
\end{figure*}

\section{Qualitative Results on Multiple Person Segmentation \label{sec:app_samples}}
In the following, we show some results produced by our model on multiple person segmentation.

\begin{figure}[h]
\begin{centering}
\begin{tabular}{cccc}

Input & RIS & RIS+CRF & Ground Truth \tabularnewline

\includegraphics[width=0.24\columnwidth]{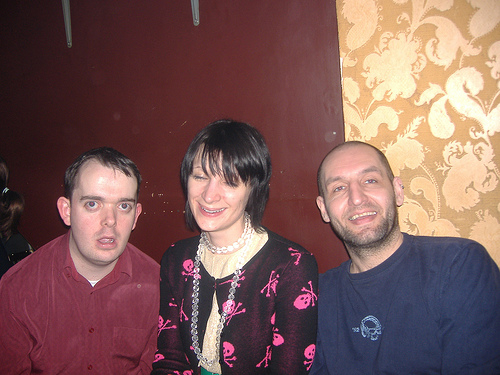} & \includegraphics[width=0.24\columnwidth]{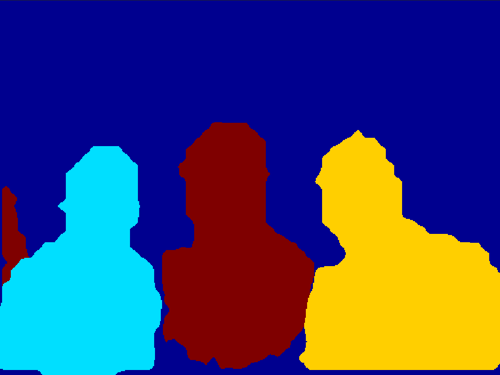} & \includegraphics[width=0.24\columnwidth]{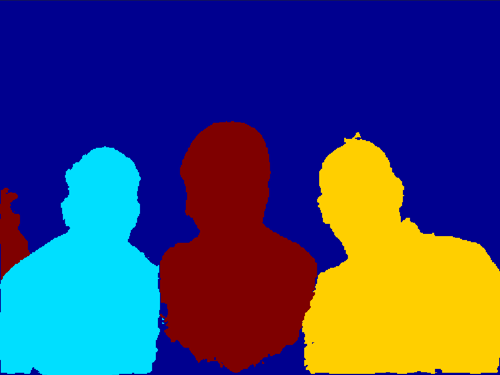} & \includegraphics[width=0.24\columnwidth]{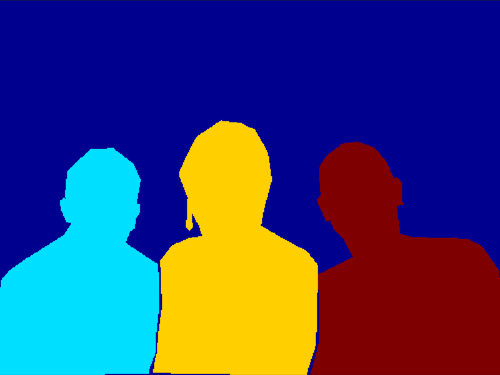}  
\tabularnewline

\includegraphics[width=0.24\columnwidth]{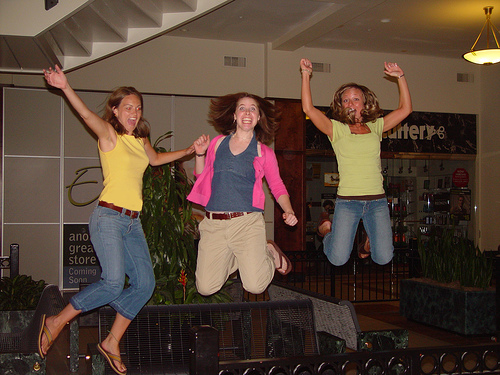} & \includegraphics[width=0.24\columnwidth]{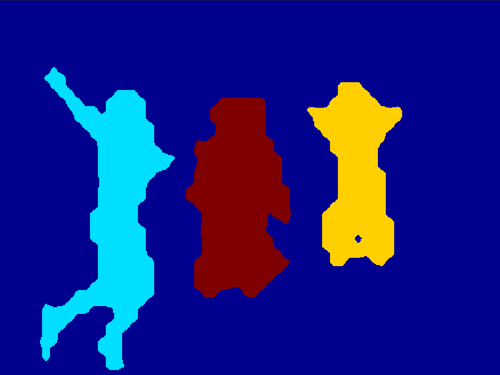} & \includegraphics[width=0.24\columnwidth]{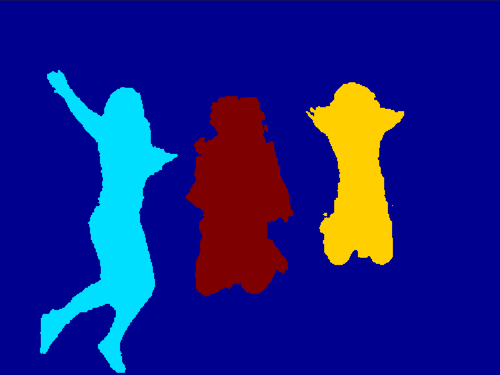} & \includegraphics[width=0.24\columnwidth]{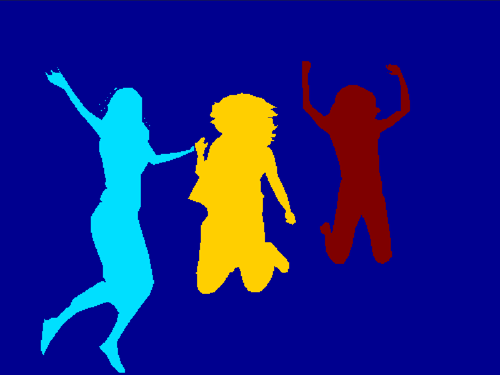}  
\tabularnewline

\includegraphics[width=0.24\columnwidth]{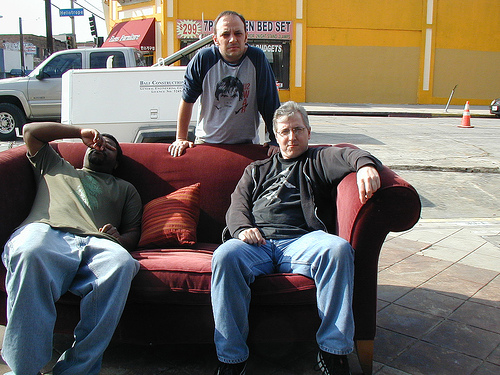} & \includegraphics[width=0.24\columnwidth]{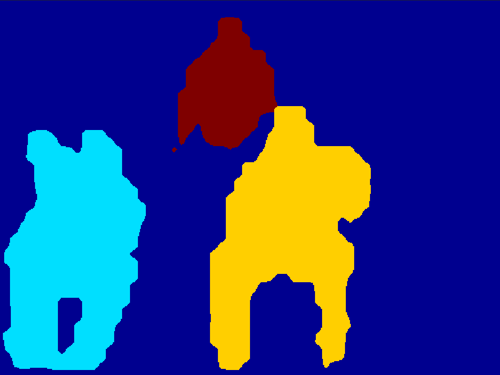} & \includegraphics[width=0.24\columnwidth]{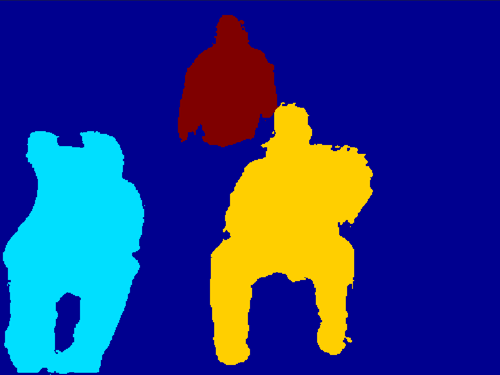} & \includegraphics[width=0.24\columnwidth]{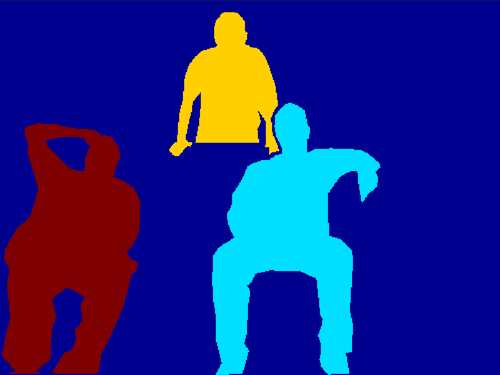}  
\tabularnewline

\includegraphics[width=0.24\columnwidth]{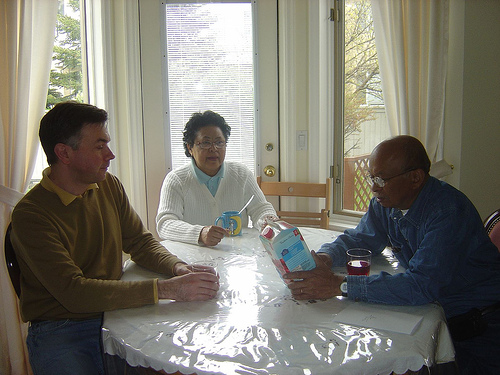} & \includegraphics[width=0.24\columnwidth]{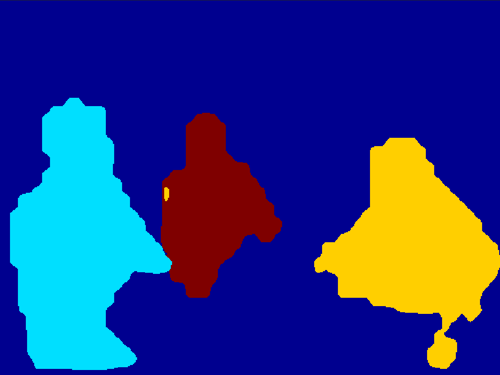} & \includegraphics[width=0.24\columnwidth]{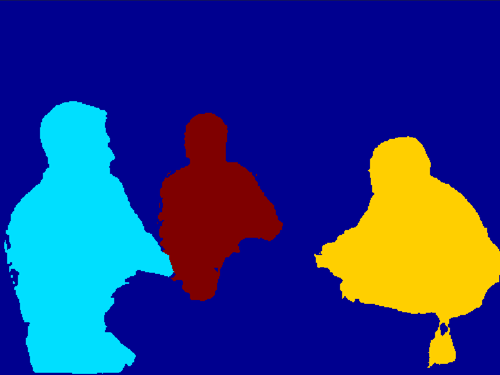} & \includegraphics[width=0.24\columnwidth]{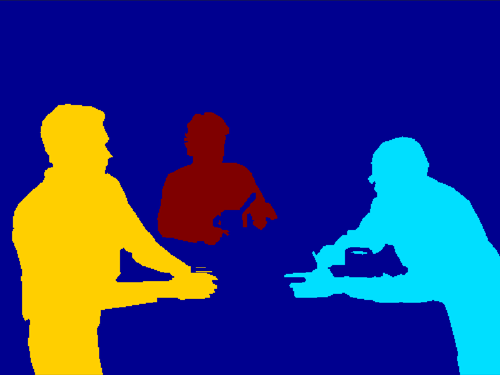}  
\tabularnewline

\includegraphics[width=0.24\columnwidth]{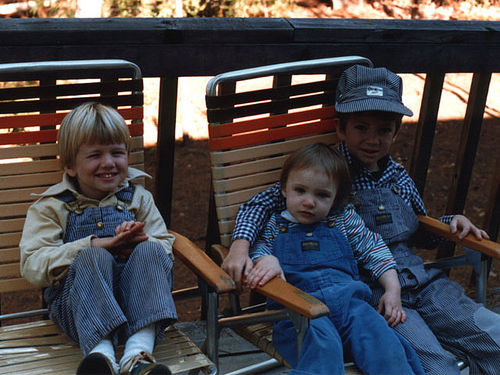} & \includegraphics[width=0.24\columnwidth]{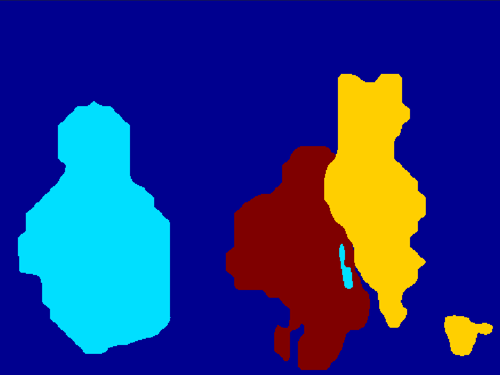} & \includegraphics[width=0.24\columnwidth]{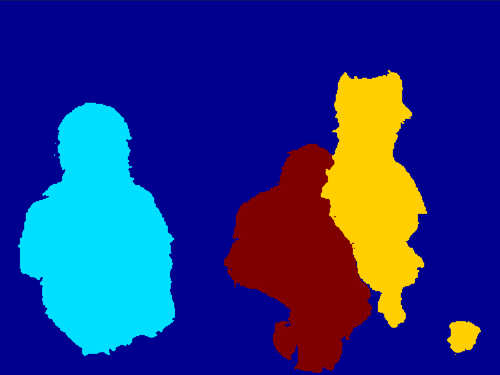} & \includegraphics[width=0.24\columnwidth]{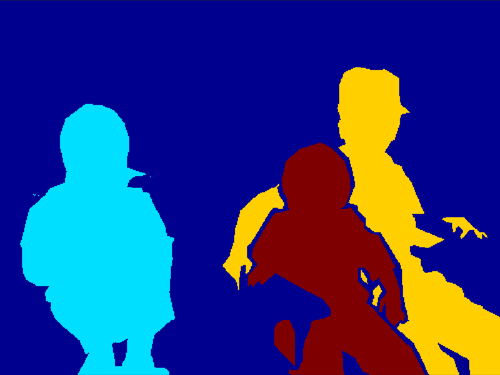}  
\tabularnewline

\includegraphics[width=0.24\columnwidth]{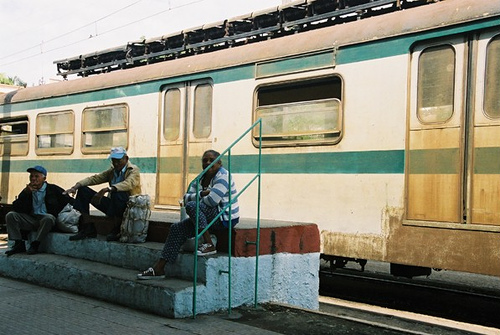} & \includegraphics[width=0.24\columnwidth]{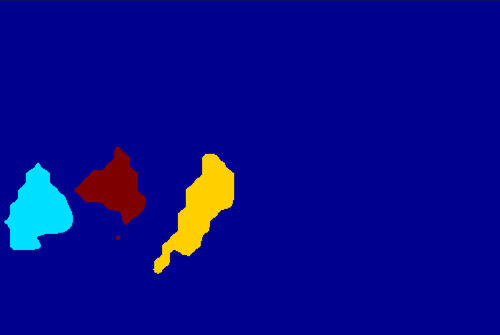} & \includegraphics[width=0.24\columnwidth]{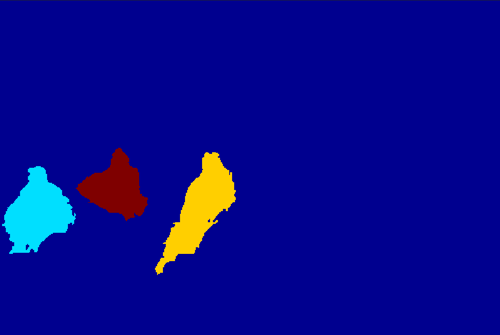} & \includegraphics[width=0.24\columnwidth]{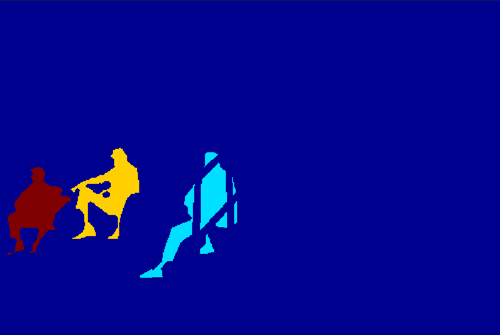}  
\tabularnewline

\includegraphics[width=0.24\columnwidth]{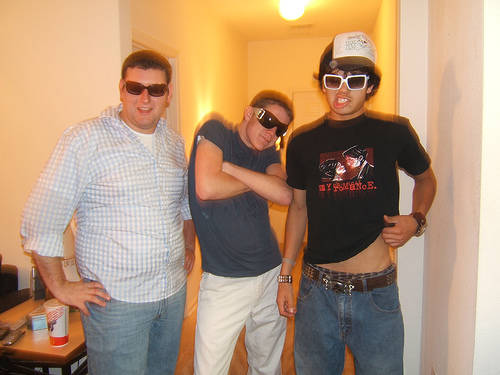} & \includegraphics[width=0.24\columnwidth]{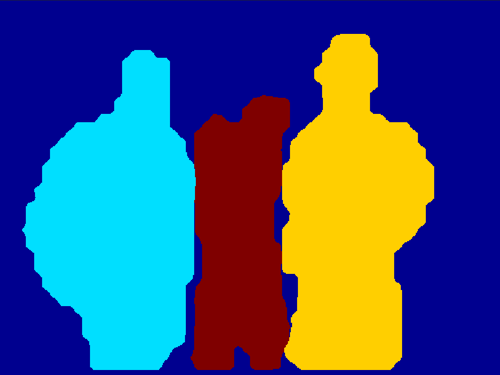} & \includegraphics[width=0.24\columnwidth]{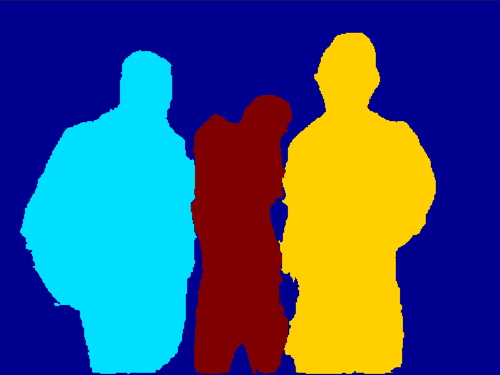} & \includegraphics[width=0.24\columnwidth]{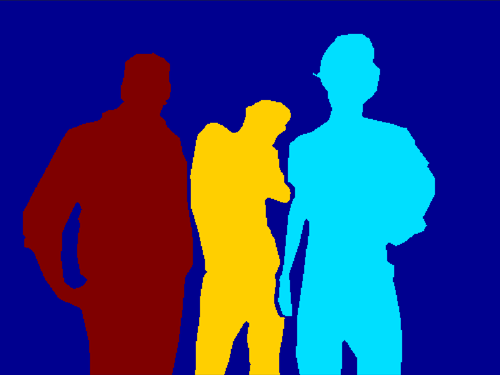}  
\tabularnewline

\end{tabular}

\par\end{centering}

\caption{\label{fig:3instances} Typical good inferences made by our model, where the input image contains $3$ subjects. Input images taken from the VOC Pascal 2012 validation dataset. Best viewed in colour.}
\end{figure}

In Fig. \ref{fig:3instances}, we show the outputs of our model (with and without CRF post-processing), with some input images containing $3$ subjects as ground truth. Note that the difference in colour (order) between ground truth instances and predictions is irrelevant. In these images the main errors are due to the difficulty to segment extremities (e.g. images in the second row).

\begin{figure}[h]
\begin{centering}
\begin{tabular}{cccc}

Input & RIS & RIS+CRF & Ground Truth \tabularnewline

\includegraphics[width=0.24\columnwidth]{2007_003020_rgb} & \includegraphics[width=0.24\columnwidth]{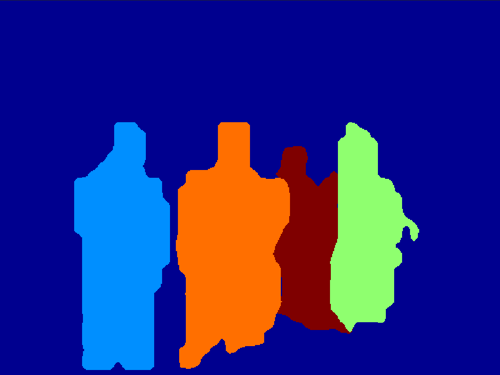} & \includegraphics[width=0.24\columnwidth]{2007_003020_crf} & \includegraphics[width=0.24\columnwidth]{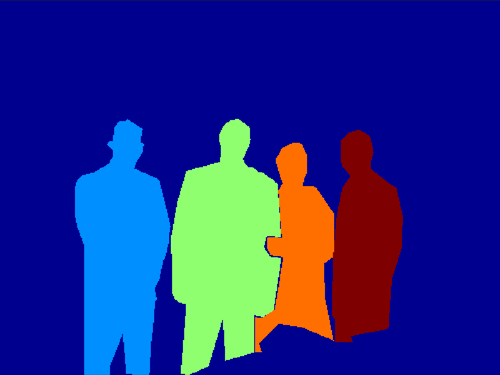}  
\tabularnewline

\includegraphics[width=0.24\columnwidth]{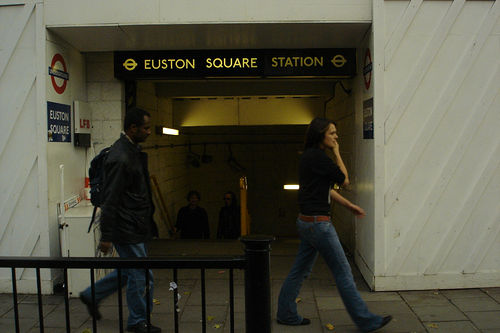} & \includegraphics[width=0.24\columnwidth]{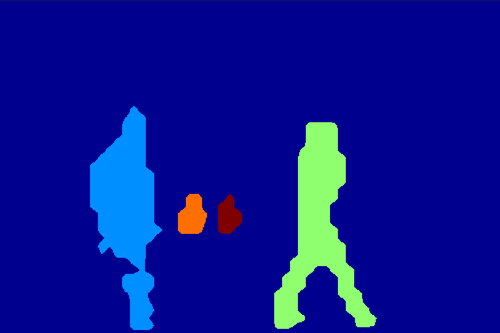} & \includegraphics[width=0.24\columnwidth]{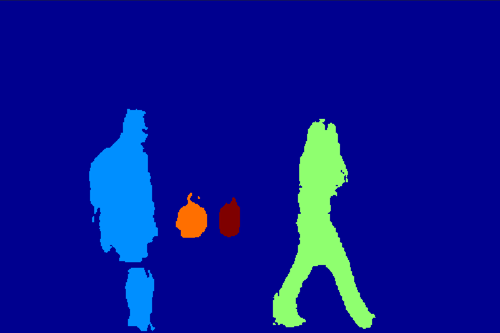} & \includegraphics[width=0.24\columnwidth]{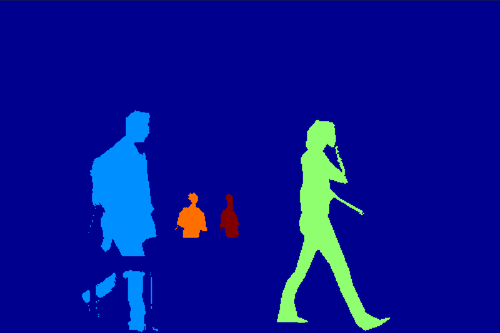}  
\tabularnewline

\includegraphics[width=0.24\columnwidth]{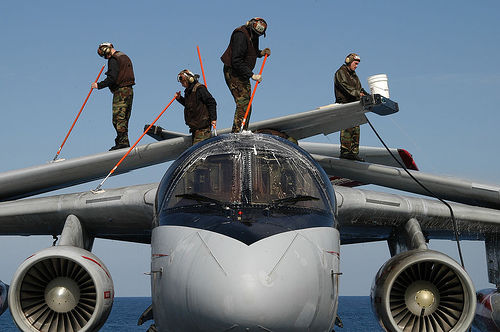} & \includegraphics[width=0.24\columnwidth]{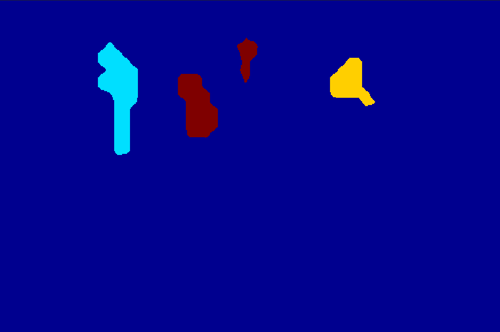} & \includegraphics[width=0.24\columnwidth]{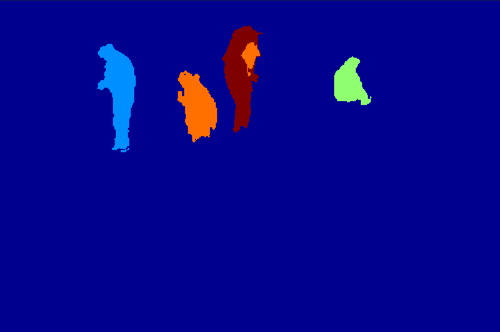} & \includegraphics[width=0.24\columnwidth]{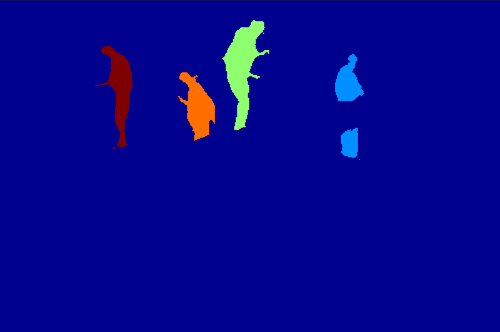}  
\tabularnewline

\includegraphics[width=0.24\columnwidth]{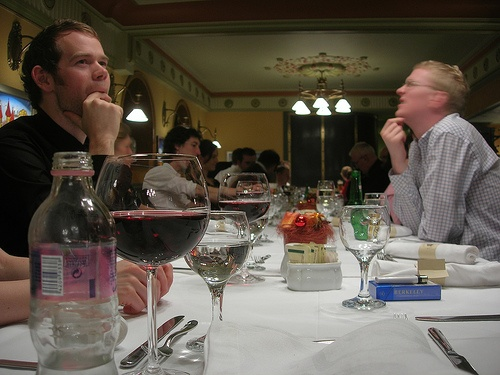} & \includegraphics[width=0.24\columnwidth]{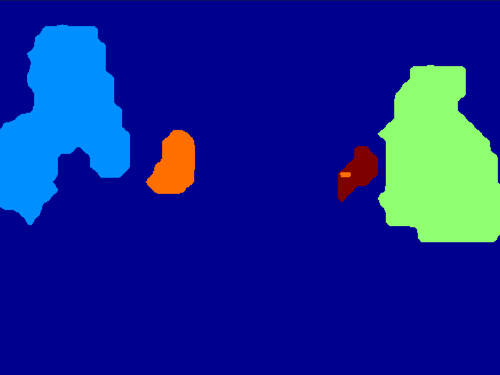} & \includegraphics[width=0.24\columnwidth]{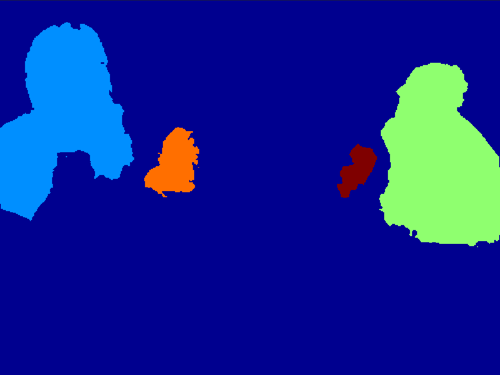} & \includegraphics[width=0.24\columnwidth]{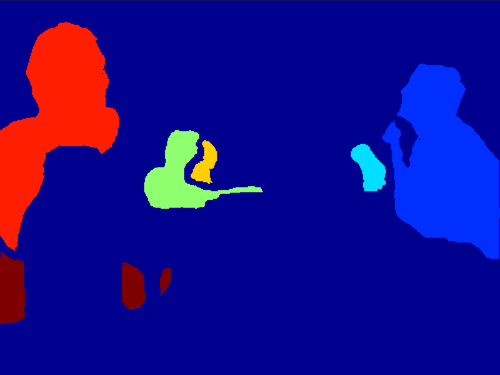}  
\tabularnewline

\includegraphics[width=0.24\columnwidth]{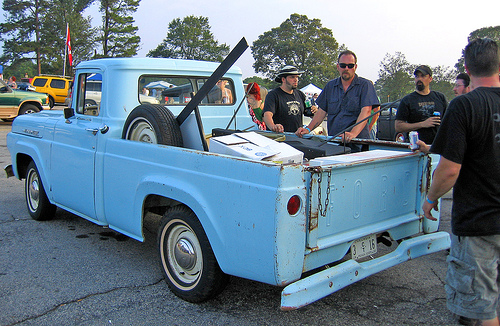} & \includegraphics[width=0.24\columnwidth]{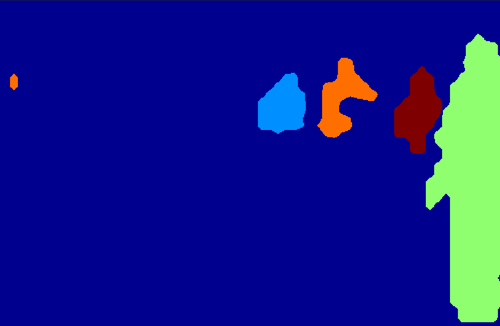} & \includegraphics[width=0.24\columnwidth]{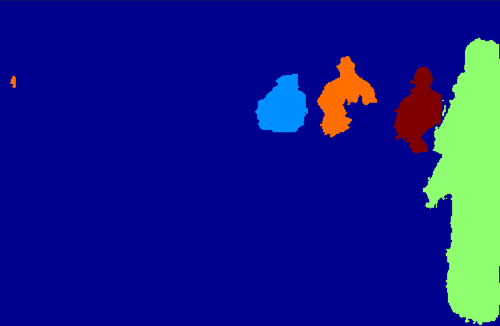} & \includegraphics[width=0.24\columnwidth]{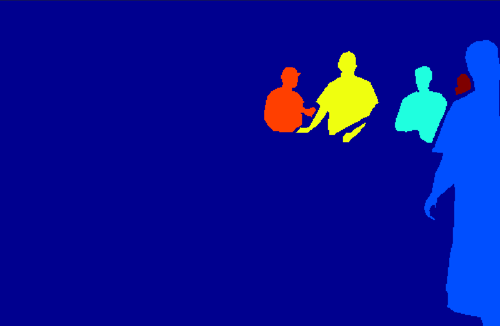}  
\tabularnewline

\includegraphics[width=0.24\columnwidth]{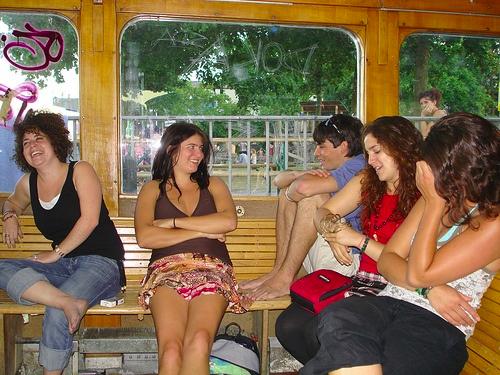} & \includegraphics[width=0.24\columnwidth]{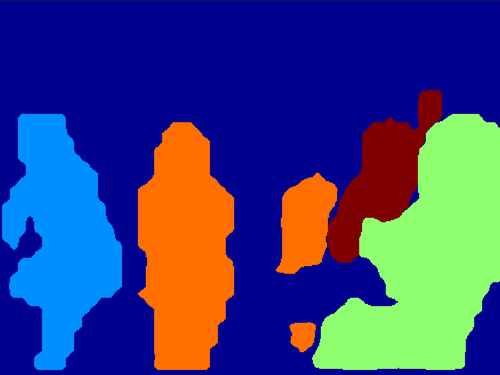} & \includegraphics[width=0.24\columnwidth]{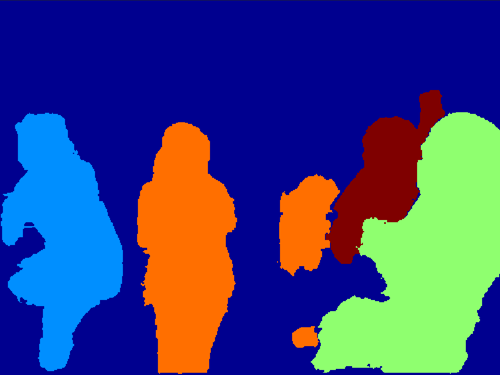} & \includegraphics[width=0.24\columnwidth]{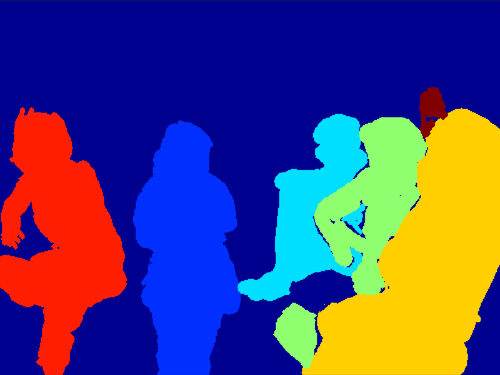}  
\tabularnewline

\includegraphics[width=0.24\columnwidth]{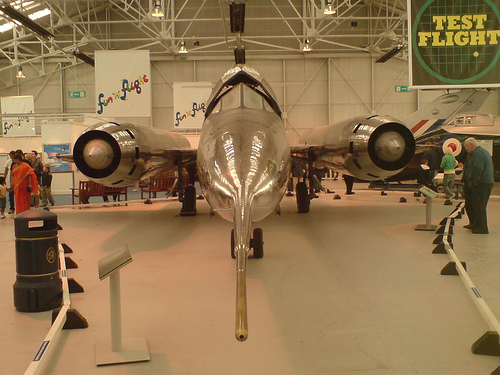} & \includegraphics[width=0.24\columnwidth]{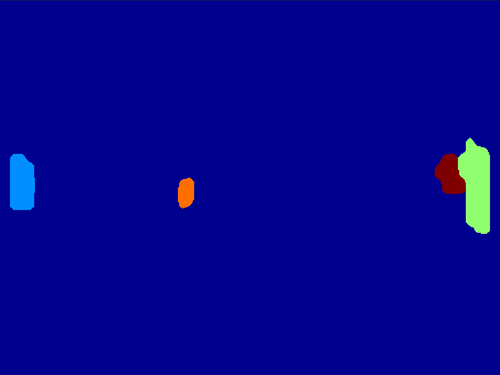} & \includegraphics[width=0.24\columnwidth]{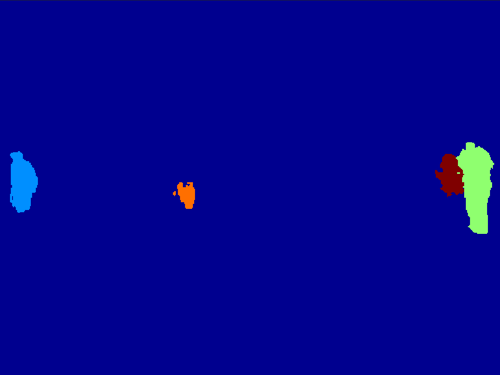} & \includegraphics[width=0.24\columnwidth]{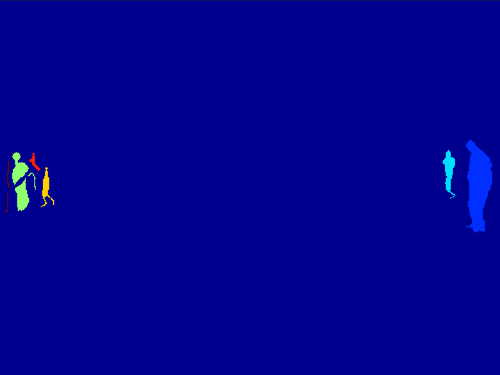}  
\tabularnewline

\end{tabular}

\par\end{centering}

\caption{\label{fig:4instances} Typical inferences made by our model, where the input image contains $4$ or more subjects. Input images taken from the VOC Pascal 2012 validation dataset. Best viewed in colour.}
\end{figure}

Similarly, in Fig. \ref{fig:4instances}, we show the prediction of our model with images that contain $4$ or more subjects. In this case we see that our model struggles when segmenting more than $4$ instances (e.g. images in the sixth row), yet it is still able to provide good inferences, even in cases when the instances (subjects) appear far away (e.g. images in the second row).

\begin{figure}[h]
\begin{centering}
\begin{tabular}{cccc}

Input & RIS & RIS+CRF & Ground Truth \tabularnewline

\includegraphics[width=0.24\columnwidth]{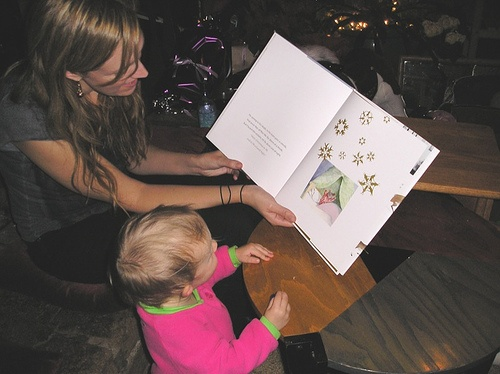} & \includegraphics[width=0.24\columnwidth]{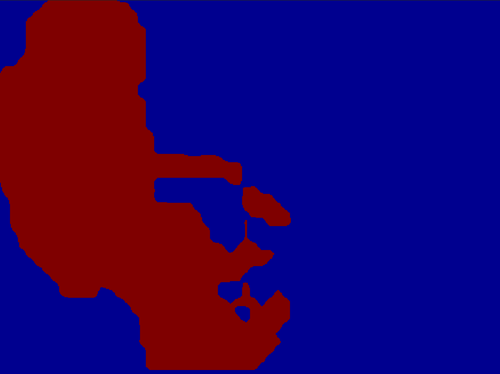} & \includegraphics[width=0.24\columnwidth]{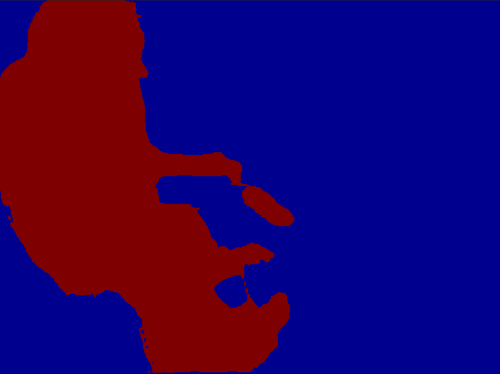} & \includegraphics[width=0.24\columnwidth]{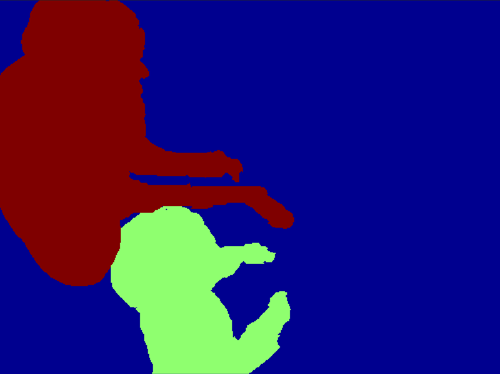}  
\tabularnewline

\includegraphics[width=0.24\columnwidth]{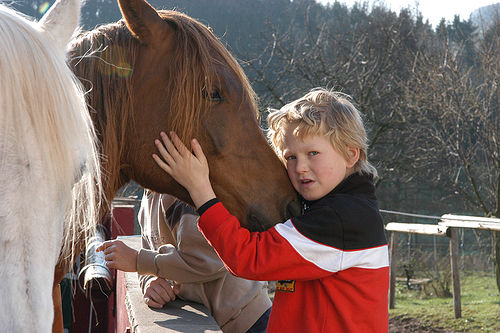} & \includegraphics[width=0.24\columnwidth]{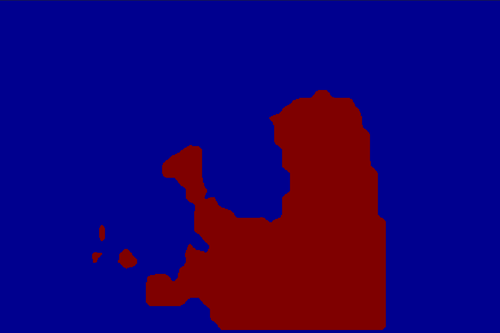} & \includegraphics[width=0.24\columnwidth]{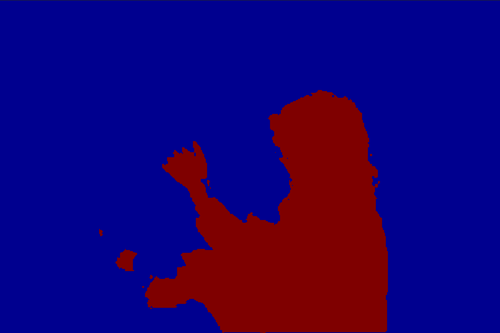} & \includegraphics[width=0.24\columnwidth]{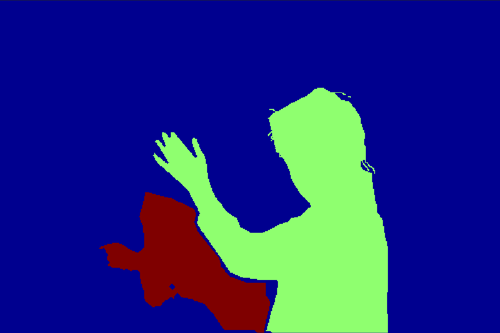}  
\tabularnewline

\end{tabular}
\par\end{centering}
\caption{\label{fig:fail1} Samples of failure cases: joining instances. Input images taken from the VOC Pascal 2012 validation dataset. Best viewed in colour.}
\end{figure}

\begin{figure}[h]
\begin{centering}
\begin{tabular}{cccc}

Input & RIS & RIS+CRF & Ground Truth \tabularnewline

\includegraphics[width=0.24\columnwidth]{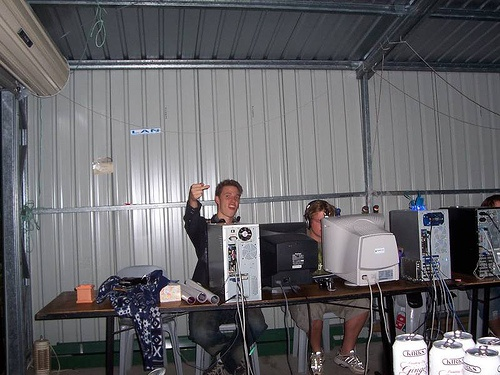} & \includegraphics[width=0.24\columnwidth]{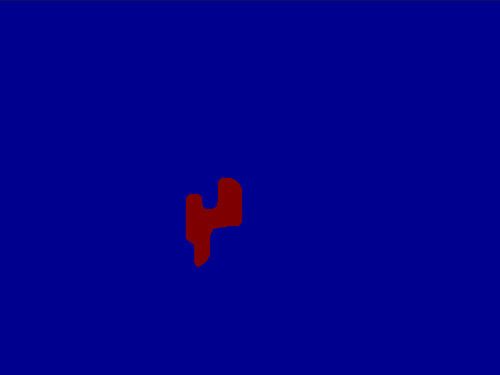} & \includegraphics[width=0.24\columnwidth]{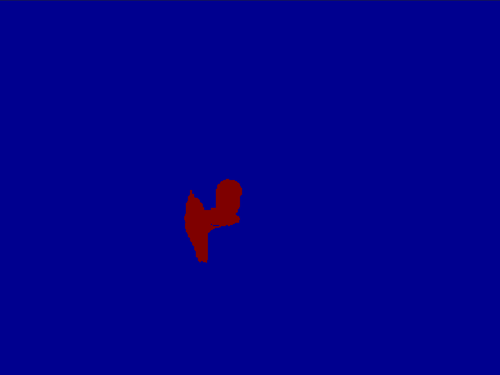} & \includegraphics[width=0.24\columnwidth]{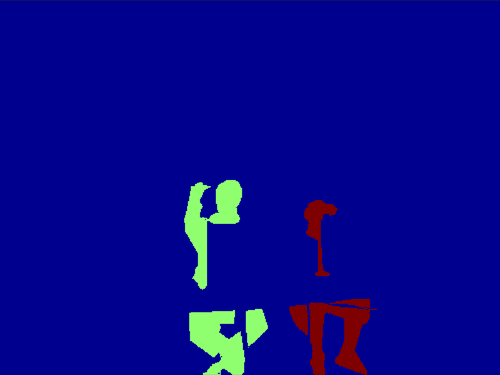}  
\tabularnewline

\includegraphics[width=0.24\columnwidth]{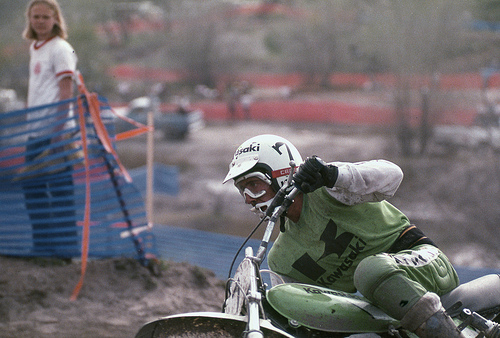} & \includegraphics[width=0.24\columnwidth]{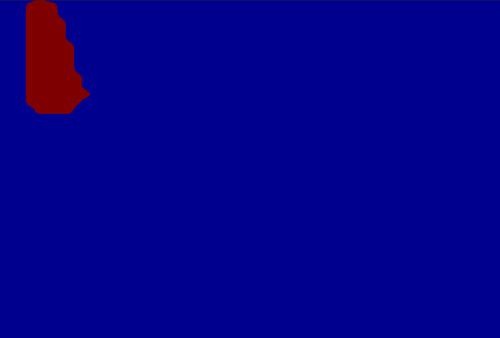} & \includegraphics[width=0.24\columnwidth]{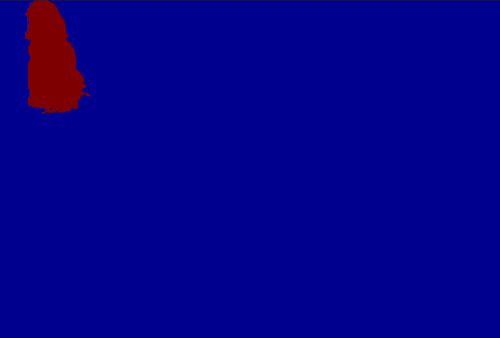} & \includegraphics[width=0.24\columnwidth]{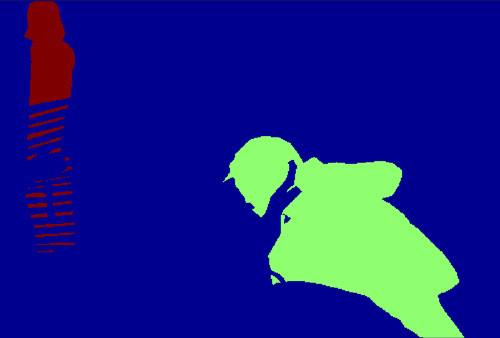}  
\tabularnewline

\end{tabular}
\par\end{centering}
\caption{\label{fig:fail2} Samples of failure cases: missing instances. Input images taken from the VOC Pascal 2012 validation dataset. Best viewed in colour.}
\end{figure}

\begin{figure}[h]
\begin{centering}
\begin{tabular}{cccc}

Input & RIS & RIS+CRF & Ground Truth \tabularnewline

\includegraphics[width=0.24\columnwidth]{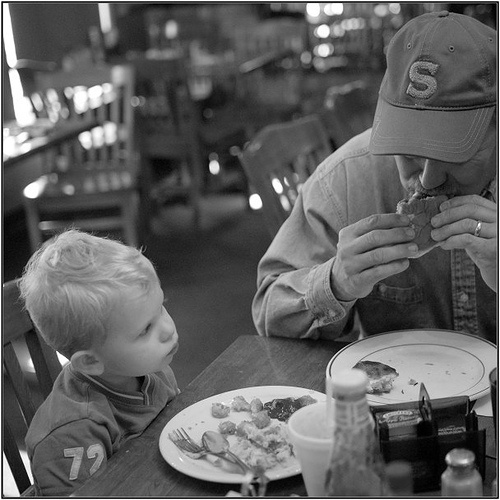} & \includegraphics[width=0.24\columnwidth]{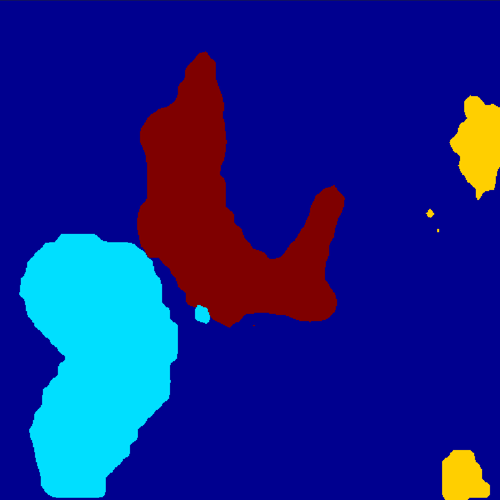} & \includegraphics[width=0.24\columnwidth]{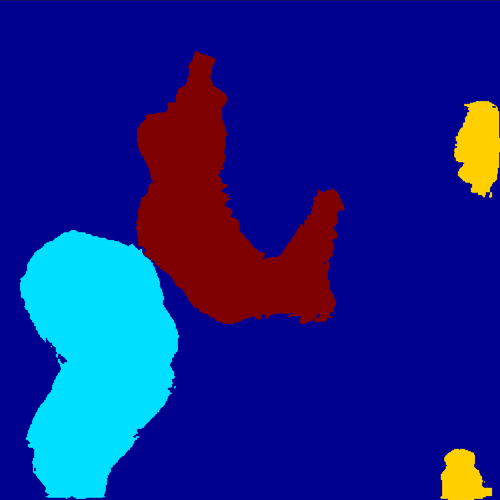} & \includegraphics[width=0.24\columnwidth]{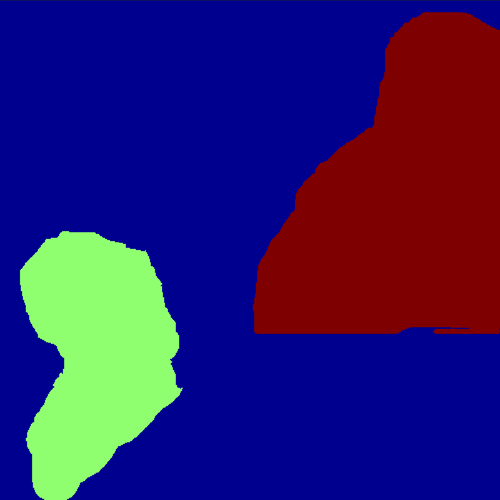}  
\tabularnewline

\includegraphics[width=0.24\columnwidth]{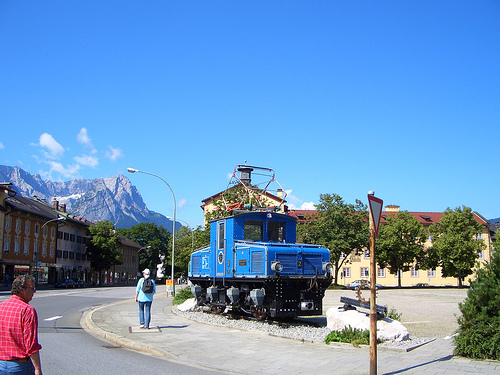} & \includegraphics[width=0.24\columnwidth]{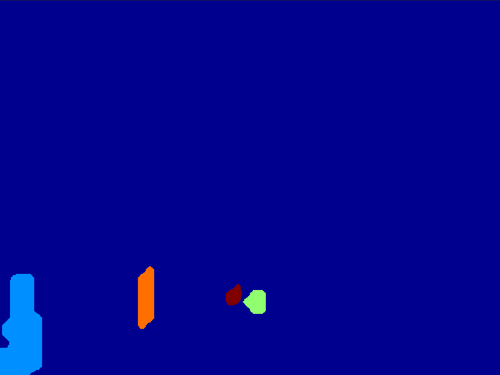} & \includegraphics[width=0.24\columnwidth]{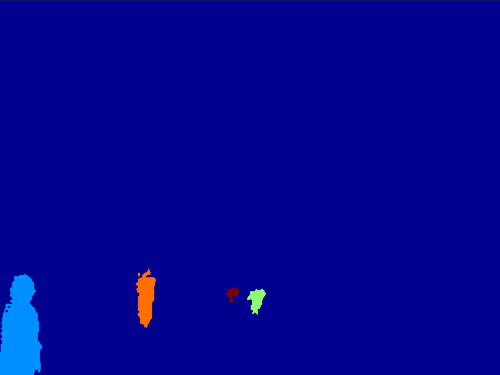} & \includegraphics[width=0.24\columnwidth]{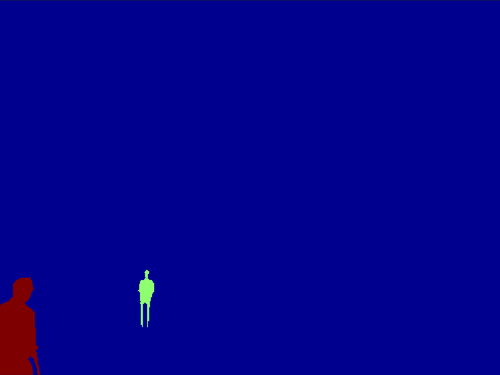}  
\tabularnewline

\end{tabular}
\par\end{centering}
\caption{\label{fig:fail3} Samples of failure cases: hallucination. Input images taken from the VOC Pascal 2012 validation dataset. Best viewed in colour.}
\end{figure}

Finally, we have collected a set of failure cases. We have divided these in three classes which cover almost the entire range of failures we have seen. Firstly, our model might segment two instances as if they were only one. Examples of this are shown in Fig. \ref{fig:fail1}. Secondly, our model sometimes misses an instance entirely, such as shown in the images in Fig. \ref{fig:fail2}. Finally, we have seen that our model might hallucinate, that is, it might predict instances where there are none, see for example the images shown in Fig. \ref{fig:fail3}.

\section{Implementation of the Loss Function \label{sec:app_loss}} 
In Algorithm \ref{alg:loss} we provide a detailed implementation of the forward and backward propagation of the loss function in eq. (\ref{eq:loss}).

\begin{algorithm}[h]
\begin{algorithmic}
\STATE \textbf{Input}: ground truth $\mathbf{Y}=\left\{ \mathbf{Y_{1}},\ldots,\mathbf{Y_{n}}\right\} $,
predicted masks $\mathbf{\hat{Y}}=\left\{ \mathbf{\hat{Y}_{1}},\ldots,\mathbf{\hat{Y}_{\hat{n}}}\right\} $,
and predicted scores $\mathbf{s}=\left\{ s_{1},\ldots,s_{\hat{n}}\right\} $
\STATE \textbf{Hyperparameters}: $\lambda$
\STATE \textbf{Output}: cost $c$, and gradients $\delta\mathbf{\hat{Y}}$, $\delta\mathbf{s}$
\STATE \textbf{Initialization}: $M\in[0,1]^{\hat{n}\times n}$, $\delta\mathbf{\hat{Y}}=\left\{ \delta\mathbf{\hat{Y}_{1}},\ldots,\delta\mathbf{\hat{Y}_{\hat{n}}}\right\} $, $\delta\mathbf{s}=\left\{ \delta s_{1},\ldots,\delta s_{\hat{n}}\right\} $
, $c=0$

\STATE
\STATE \textit{Forward:}
\STATE Fill $M$, so that $M_{i,j}=f_{\rm IoU}(\mathbf{Y_{i}},\mathbf{\hat{Y}_{j}})$.
\STATE ${\rm matching}={\rm Hungarian}(M)$
\FOR {$t=1\ldots n$}
\STATE $c=c-M_{t,{\rm matching}(t)}+\lambda f_{\rm BCE}(1,s_{t})$
\ENDFOR
\FOR {$t=n+1\ldots\hat{n}$}
\STATE $c=c+\lambda f_{\rm BCE}(0,s_{t})$
\ENDFOR
\STATE 
\STATE \textit{Backward:}
\FOR {$t=1\ldots\hat{n}$}
\IF {$t\leq n$}
\STATE $\delta\mathbf{\hat{Y}_{t}}=f'_{IoU}(\mathbf{Y_{{\rm matching}(t)}},\mathbf{\hat{Y}_{t}})$
\STATE $\delta s_{t}=\lambda f'_{BCE}(1,s_{t})$
\ELSE
\STATE $\delta s_{t}=\lambda f'_{BCE}(0,s_{t})$
\ENDIF 
\ENDFOR

\end{algorithmic}
\caption{Forward and backward propagation of the loss function in eq. (\ref{eq:loss})}
\label{alg:loss}
\end{algorithm}

The function ${\rm Hungarian}$ finds the best matching given the matrix of values as an input. The functions $f'_{IoU}$ and $f'_{BCE}$ are the derivative functions of
$f{}_{IoU}$ and $f{}_{BCE}$ with respect to their second argument.

\end{document}